%% file: bare_jrnl.tex
\documentclass[journal]{IEEEtran}
\usepackage{cite}

\ifCLASSINFOpdf
   \usepackage[pdftex]{graphicx}
\else
\fi
\usepackage{amsmath}
\usepackage{amsfonts}
\usepackage{amssymb}

\ifCLASSOPTIONcompsoc
  \usepackage[caption=false,font=normalsize,labelfont=sf,textfont=sf]{subfig}
\else
  \usepackage[caption=false,font=footnotesize]{subfig}
\fi
\usepackage{url}

\usepackage{tikz}
\usetikzlibrary{decorations.pathreplacing,decorations.markings}

\begin{document}
%
\title{Clustering of Driving Encounter Scenarios Using Connected Vehicle Trajectories}
%
%
%

\author{Wenshuo~Wang,~\IEEEmembership{Member,~IEEE,}
        Aditya~Ramesh,~
        and~Ding~Zhao
\thanks{The corresponding author is Ding Zhao.}
\thanks{W. Wang is with the Department of Mechanical Engeering, Carnegie Mellon University (CMU), Pittsburgh, PA. He was with the Department of Mechanical Engineering, University of Michigan (UM), Ann Arbor, MI, 48109. USA. e-mail: wwsbit@gmail.com.}
\thanks{A. Ramesh is with the Department of Electrical Engineering and Computer Science, University of Michigan, Ann Arbor,
	MI, 48109 USA e-mail: raaditya@umich.edu.}
\thanks{D. Zhao is with the Department of Mechanical Engeering, Carnegie Mellon University (CMU), Pittsburgh, PA. He was with the Department of Mechanical Engineering and also the Robotics Institute, University of Michigan, Ann Arbor, MI, 48109 USA e-mail: dingzhao@andrew.cmu.edu.}
}

\maketitle

\begin{abstract}
Multi-vehicle interaction behavior classification and analysis offer in-depth knowledge to make an efficient decision for autonomous vehicles. This paper aims to cluster a wide range of driving encounter scenarios based only on multi-vehicle GPS trajectories. Towards this end, we propose a generic unsupervised learning framework comprising two layers: feature representation layer and clustering layer. In the layer of feature representation, we combine the deep autoencoders with a distance-based measure to map the sequential observations of driving encounters into a computationally tractable space that allows quantifying the spatiotemporal interaction characteristics of two vehicles. The clustering algorithm is then applied to the extracted representations to gather homogeneous driving encounters into groups. Our proposed generic framework is then evaluated using 2,568 naturalistic driving encounters. Experimental results demonstrate that our proposed generic framework incorporated with unsupervised learning can cluster multi-trajectory data into distinct groups. These clustering results could benefit decision-making policy analysis and design for autonomous vehicles.
%
%
%
%
\end{abstract}

%
\IEEEpeerreviewmaketitle

\section{Introduction}
\begin{figure}[t]
	\centering
	\includegraphics[width = \linewidth]{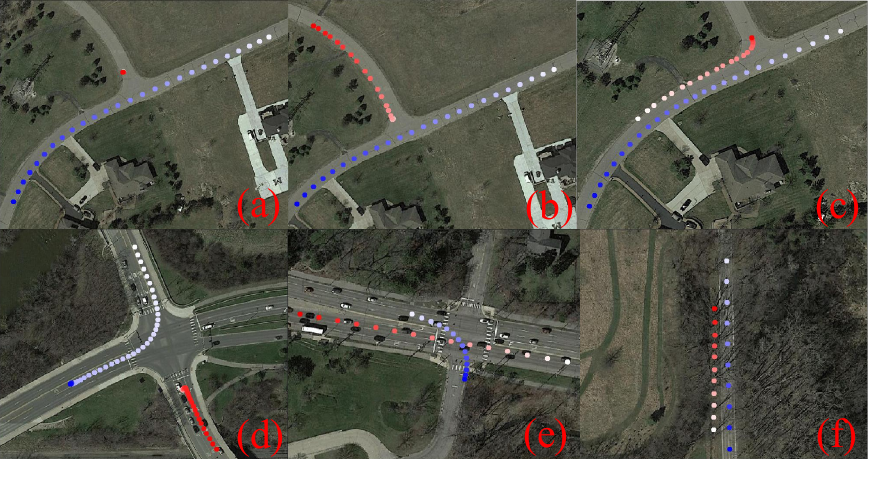}
	\caption{Examples of driving encounters at (a)-(c) T-shape intersections, (d)-(e) cross-intersections, and (f) a straight road. Thick red and blue dots are the start position of vehicles, and white dots are the end position of vehicles. }
	\label{fig:examples}
\end{figure}

%
%
%
%
\IEEEPARstart{M}{ulti-vehicle} interaction is an everyday and important driving scenario in real traffic, but very challenging for analysis and modeling because of the diversity in spatiotemporal attributes. Driving encounter in this paper is referred to as the scenario where two or multiple vehicles are spatially close to and interact with each other when driving.  In real life, autonomous vehicles should make a proper decision in various negotiation scenarios such as on highways and at intersections. An expected decision-making and control system for fully autonomous driving can adequately understand and handle all possible driving encounter scenarios to guarantee road safety and traffic efficiency. However, the diversity of encountering scenarios in real life overwhelms the human mind and insights, as shown in Fig. \ref{fig:examples}.

Multi-vehicle interaction modeling and analysis rely on many kinds of advanced communication techniques such as dedicated short-range communications (DSRC) for driving encounter data collection. The multi-vehicle interaction can be interpreted by their trajectory -- a set of positional information for moving vehicles, ordered by time. The growing use of global positioning system (GPS) receivers in equipped vehicles empowers researchers to collect large amounts of high-quality trajectory data at a low cost. Using positional trajectories with GPS data allows us to dig underlying information and visualize multi-vehicle behaviors with road context as shown in Fig. \ref{fig:examples}. Vehicle trajectories with GPS data have also been widely used for driver behavior pattern analysis and prediction\cite{deo2018would,taylor2015method,wang2017driving}, travel destination prediction\cite{besse2017destination}, anomalous driving behavior detection\cite{piciarelli2008trajectory}, eco-driving\cite{sun2015trajectory}, vehicle behavior reconstruction\cite{chen2017vehicle}. However, the diversity of driving encounters at variance with driving conditions (e.g., road, traffic, and other road users) and the flood of high-dimensional traffic data overwhelms human insight and analysis\cite{appenzeller2017scientists}, thereby limiting to make full use of these trajectory data. Fig. \ref{fig:examples} presents some typical driving encounters consisting of two vehicles at different intersections. Therefore, clustering driving encounters into distinct groups enables us to identify the common underlying points, encourages a vibrant design of space and functionality for synthetic traffic systems, and highlights a compelling need for the integrative analysis of the interaction between human drivers or between the human-driven vehicle and autonomous vehicles.

Many clustering algorithms and data mining techniques have been implemented to vehicle trajectory analysis and associated applications, as reviewed in \cite{yuan2017review,feng2016survey};
however, most of them aim to discover the group of similar single trajectory, rather than the group of similar multi-vehicle trajectories in driving encounter scenarios.
For example, Besse \textit{et al}.\cite{besse2016review}  proposed a distance-based algorithm to learn representations of each single vehicle trajectory and then applied hierarchical clustering to the representations of all vehicles for categorization. The authors in \cite{besse2017destination} also developed a two-step procedure to predict a trip's destination using a density-based clustering of destination and the initial part of trajectories. Yao  \textit{et al}.\cite{yao2018learning} used a sliding window to extract a set of moving behavior features that capture space and time-invariant characteristics of the trajectories. Zhao \textit{et al}. \cite{zhao2017road,yao2017road} extracted lane change behavior according to the vehicle's raw movement trajectory and conducted the scene-based analysis. In order to obtain traffic patterns for traffic surveillance, Choong \textit{et al}. \cite{choong2017modeling} implemented the longest common subsequence (LCSS) to measure similarity levels of trajectories as features and then clustered vehicle trajectories into groups using $ k $-means and fuzzy $ c $-means. Some research \cite{zhao2017trajectory,feng2016survey,wang2017automatic,zhan2017citywide,kim2015spatial} also applied clustering algorithms to detect traffic information such as traffic hotspots, traffic patterns and traffic volume based on vehicle GPS trajectories. 

However, all these methods above are unsuitable for clustering a set of multi-vehicle trajectories because of their limited ability to represent high-dimensional sequential observations. Investigating multi-vehicle interaction patterns (driving encounter scenarios) can benefit insights about how human drivers make decisions when negotiating with others, thereby providing external knowledge for self-driving. For instance, classifying cross-negotiation behaviors at unsignalized intersections can offer researchers detailed and applicable information to design rule-based decision-making systems for autonomous vehicles that can harmoniously negotiate with surrounding road users\cite{galceran2017multipolicy}. Towards this goal, many researchers used vehicle-to-vehicle (V2V) trajectories in the empirically predefined specific scenarios such as changing lanes. Although achieved remarkable progress has been made, the scenarios they concerned are limited in diversity, which sharply limits the potential applications. Therefore, categorizing similar driving encounters into groups can benefit insight into each kind of encounter scenario and offer opportunities to test the existing algorithms. For example, Pokorny \textit{et al}.\cite{pokorny2016topological} proposed a topological trajectory clustering by considering the relative persistent homology between motion trajectories. Our previous work \cite{li2018clustering} directly applied a common autoencoder to extract features of characterizing driving encounters for clustering, but the learned representations were not interpretable. To the best of our knowledge, great efforts on tackling a bunch of single trajectories have been made by utilizing off-the-shelf algorithms, but no one efficient approach is suitable for clustering a set of multi-vehicle trajectories.  

In this paper, we primarily focus on the driving encounter scenario within two vehicles engaged and clustering their GPS trajectories into distinct groups. 
The main contributions of this paper are threefold.
\begin{itemize}
	\item Proposing a generic two-layer framework to cluster the driving encounter scenarios represented by multi-vehicle GPS trajectories. 
	\item Developing different unsupervised learning methods to encode driving encounters into computationally tractable measure space using deep autoencoders, distance-based measure, and their combinations.
    \item Validating our developed two-layer framework using different algorithms based on 2,568 naturalistic driving encounters. 
\end{itemize}

The remainder of this paper is organized as follows. Section II presents the framework of clustering driving encounters. Section III details the approaches of extracting representatives. Section IV presents the clustering method and performance metrics. Section V introduces the experimental procedure and data collection. Section VI analyzes and discusses the experimental results, followed by a conclusion and future work in Section VII.

\begin{figure}[t]
	\centering
	\includegraphics[width = 0.9\linewidth]{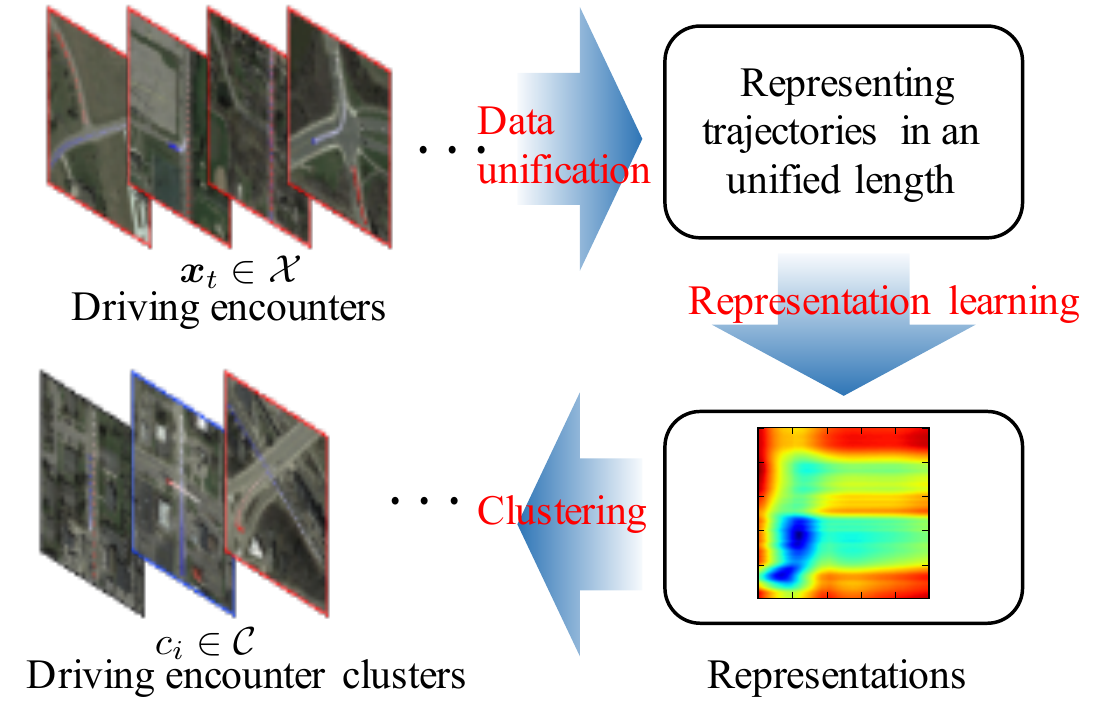}
	\caption{Our proposed framework of clustering driving encounters.}
	\label{fig:framework}
\end{figure}
\section{Clustering Framework}
In this section, we shall discuss the generic framework of clustering driving encounters, as shown in Fig. \ref{fig:framework}. This framework can be realized by three steps: driving encounter unification, representation extraction, and clustering. Before detailing the framework, we first introduce the mathematical formulation of driving encounters.

\subsection{Driving Encounter}

\begin{figure}
	\centering
	\resizebox{0.95\linewidth}{!}{\input{figures/drivingencounter.tex}}
	\caption{Illustration of one driving encounter with two vehicle trajectories.}
	\label{fig:drivingencounter}
\end{figure}
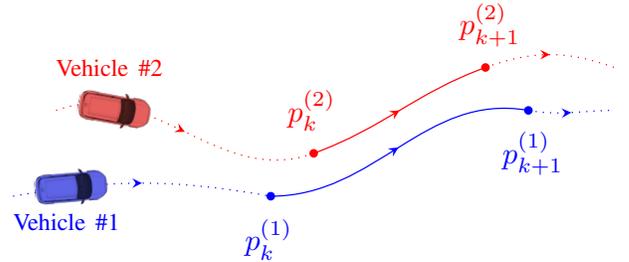

Two vehicle trajectories with the same length represent a driving encounter, where the vehicle trajectories could be continuous or discrete. In our case, we will consider the discrete trajectories, defined as follows. For a single driving encounter, we obtained the discrete observations 

\begin{equation}
\boldsymbol{x} = \left\lbrace  (p^{(1)}_{1}, p^{(2)}_{1}, t_{1}), \cdots, (p^{(1)}_{k}, p^{(2)}_{k}, t_{k}), \cdots, (p^{(1)}_{K}, p^{(2)}_{K}, t_{K}) \right\rbrace 
\end{equation}
where $ p^{(1)}_{k} \in \mathbb{R}^{2} $  and $ p^{(2)}_{k} \in \mathbb{R}^{2} $ are the position of two vehicles in the driving encounter at discrete time $ t_{k} $ (Fig. \ref{fig:drivingencounter}), $ K \in \mathbb{N} $ is the length of the driving encounter $ \boldsymbol{x} $. In particular, we define $ \boldsymbol{X} = (\boldsymbol{x}_{1}, \boldsymbol{x}_{2}, \cdots, \boldsymbol{x}_{N})$ as a set of driving encounters with $ N $ samples.

\subsection{Driving Encounter Unification}
For clustering algorithms, all the input samples or features should be in the identical dimension; therefore the raw driving encounter GPS trajectories are required to scale to a preset length. For two arbitrary driving encounters in real traffic, their lengths could be significantly different from each other. In order to make clustering algorithms practically tractable, we need to scale the driving encounter data into an identical length with little loss of information. Many ripe approaches to unifying trajectories have been developed, see review literature \cite{bian2018survey}, for example,
\begin{itemize}
	\item Trajectory transformation algorithms, which project the trajectories into a different structured space with a fixed parameter, for example, linear transformation, curving fitting (e.g., uniform cubic B-spline curve), discrete Fourier transformation, etc.
	\item Re-sampling methods, which choose trajectory points by a sampling rule to unify trajectory lengths but usually result in some information loss.
	\item Trajectory substitute, which utilizes the essential components of trajectories (termed as sub-trajectories) to represent hidden information of original trajectory data.
	\item Points of interest, which is flexible and preferred when research focuses on some specific regions of surveillance (termed as points of interest) rather than the points outside the regions.
	\item Scale-invariant features, which have been widely used to extract more robust and representative features (e.g., histograms of oriented gradients and histograms of optical flow) from image frames rather than the positional trajectories.
\end{itemize}
Though the last three approaches could perform very well for a single trajectory, it is nontrivial to directly apply them to the pair of multi-vehicle trajectories in driving encounters. Hence, we prefer to use the re-sampling method with interpolation processes as it is very flexible to operate. More calculation details see \cite{bian2018survey}. The unification task can be completed using the command \texttt{scipy.interpolate} in Python.

\subsection{Representation Learning}
Straightforwardly implementing the common clustering algorithms to the sequential observations of driving encounters is practically intractable because it is meaningless to compute the center of multi-vehicle trajectories. In order to solve this problem, we transfer the multi-vehicle trajectories into a specified space and then select associated representations to characterize the interaction of vehicles in driving encounters. Manipulating on representations makes it calculable to maximize the similarity of samples within and between clusters. Many different approaches have been developed to learn representations of capturing the temporal and spatial relations of multi-vehicle trajectories, for example, by measuring the distance between two trajectories \cite{besse2016review,besse2017destination} or by using neural networks to learn representations of individual trajectories\cite{yao2018learning,yao2017trajectory}. We shall introduce three kinds of approaches to learn representations in Section III.

\subsection{Clustering}
These learned representations enable us to make full use of the off-the-shelf clustering algorithms (e.g., $k$-means) to gather driving encounters into groups. The clustering performance can also be evaluated based on these representations. More details are shown in Section IV.

\section{Representation Learning}
Clustering similar driving encounters into groups requires to extract the features capable of capturing their primary characteristics. Unlike extracting features of images with specific and explicit labels, we have limited prior knowledge about the feature space of driving encounters. Fortunately, many approaches to learning representations of time-series trajectories have been developed to study sequential data. In this paper, we mainly focus on the types of multi-vehicle dynamic interactions in spatial and temporal spaces. We shall introduce three ways to achieve this: deep autoencoders, distance-based measure, and shape-based measure. The limitations and advantages of these methods are then discussed.
%
%

\begin{figure}[t]
	\centering
	\includegraphics[width = \linewidth]{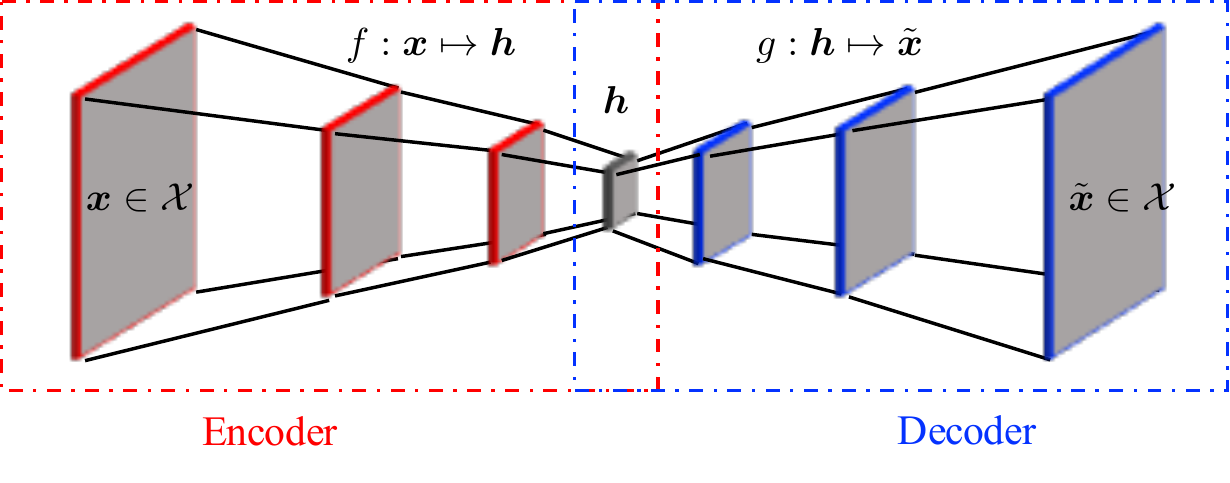}
	\caption{Illustration of representation learning through the deep neural network-based autoencoder, with an encoder $ f $ and a decoder $ g $ to learn hidden representations $ \boldsymbol{h} $ given observations $ \boldsymbol{x} \in \mathcal{X} $.}
	\label{fig:autoencoder}
\end{figure}

\subsection{Deep Autoencoders}
Our goal is to characterize a given driving encounter by finding a representation from which the driving encounter can be completely reconstructed. The autoencoder, which could generate a sample from the learned representation as close as possible to its original input, has been widely used to dig underlying information of time-series data. An autoencoder is a specific neural network (NN) consisting of encoder and decoder (as shown in Fig. \ref{fig:autoencoder}) that attempts to copy its input $ \boldsymbol{x} $ to its output $ \tilde{\boldsymbol{x}} $, where the hidden layer $ \boldsymbol{h} $ describes the code to represent the input $ \boldsymbol{x} $. By training a model to minimize the difference between $ \boldsymbol{x} $ and $ \tilde{\boldsymbol{x}} $, the hidden layer $ \boldsymbol{h} $ can capture the underlying characteristics of this driving encounter. Autoencoders vary in the architecture of the encoder and decoder as a consequence of the kind of input data being supplied. For example, a structure of long-short term memory (LSTM) generates an LSTM autoencoder, and a structure of convolutional NN (CNN) creates a convolutional autoencoder. In general, the encoder and the decoder can be formulated as follows.

\begin{itemize}
	\item \textit{Encoder}: The encoder (red dash box in Fig. \ref{fig:autoencoder}) is a multilayer neural networks of mapping an input $ \boldsymbol{x}_{(i)} $ at the $ i^{\text{th}} $-layer into $ \boldsymbol{x}_{(i+1)} $ at the $ (i+1)^{\text{th}} $-layer by
	
\begin{equation}\label{eq:TrNN}
\boldsymbol{x}_{(i+1)} = \boldsymbol{w}^{\top}_{(i)} \boldsymbol{x}_{(i)} + \boldsymbol{b}_{(i)}
\end{equation} 
where $ \boldsymbol{w}_{(i)} $ is the weight matrix for transferring data from the $i^{\text{th}}$-layer to $(i+1)^{\text{th}}$-layer, $ \boldsymbol{b}_{(i)} $ is the bias vector. Note that in Fig. \ref{fig:autoencoder} the middle layer is the representation $ \boldsymbol{h} $ of driving encounters. 
	\item \textit{Decoder}: The hidden representation $ \boldsymbol{h} $ from the encoder is then mapped back to $ \tilde{\boldsymbol{x}} $ through a symmetric multilayer network.
\end{itemize}
Subsequently, an autoencoder with a simple multilayer network can be easily derived using (\ref{eq:TrNN}) without considering the past observations, which has been thoroughly investigated in \cite{li2018clustering}. Driving behavior, however, is a dynamic process in nature which depends on past information \cite{lake2015human,nechyba1998stochastic}. Hence, we take the past observation into consideration using two advanced autoencoders: LSTM-Autoencoder (LSTMAE) and convolutional autoencoder (ConvAE).

\subsubsection{LSTMAE}
In contrast to the simple autoencoder with a multilayer perceptron, the LSTMAE \cite{hochreiter1997long} takes the advantage of the temporal information by adding an information selection function, usually as a sigmoid neural net layer, to decide what information is useful and to remember this helpful information, thus transferring the information $ \boldsymbol{x}_{(i)} $ at the $ i^{\text{th}} $-layer to the representation at the $ (i+1)^{\text{th}} $-layer. The LSTMAE comprises four basic equations to achieve this, formulated by
\begin{subequations}
	\begin{align}
	z_{(i)} & = \sigma(\boldsymbol{w}_{z}^{\top}\boldsymbol{\ell}_{(i)} + \boldsymbol{u}_{z} \boldsymbol{x}_{(i-1)} + \boldsymbol{b}_{z})\\
	y_{(i)} & = \sigma(\boldsymbol{w}_{y}^{\top}\boldsymbol{\ell}_{(i)} +  \boldsymbol{u}_{z} \boldsymbol{x}_{(i-1)} + \boldsymbol{b}_{y})\\
	\hat{\boldsymbol{x}}_{(i)} & = \tanh(\boldsymbol{w}_{l}^{\top} \boldsymbol{\ell}_{(i)}+ \boldsymbol{u}_{\ell} y_{(i)} \boldsymbol{x}_{(i-1)} + \boldsymbol{b}_{\ell})\\
   {\boldsymbol{x}}_{(i)} & = (1-z_{(i)}) \boldsymbol{x}_{(i-1)} + z_{(i)} \hat{\boldsymbol{x}}_{(i)} 
	\end{align}
\end{subequations}
where $ \sigma(\cdot) $ is the sigmoid function, $ \boldsymbol{w}_{z}, \boldsymbol{w}_{y}, \boldsymbol{w}_{\ell} $ are the weights, $ \boldsymbol{b}_{z}, \boldsymbol{b}_{y}, \boldsymbol{b}_{\ell} $ are the biases, $ \boldsymbol{x}_{(i)} $ is the output vector of the LSTM unit. Therefore, given observation $ \boldsymbol{x}_{(i-1)} $ for each LSTM unit, we can propagate the output $ \boldsymbol{x}_{(i)} $ to next layer.

\subsubsection{Convolutional Autoencoder (ConvAE)}
Sharing the same structure with LSTMAE, the ConvAE describes the layer relationship using a convolution operation, instead of using a primary multiply operator.


The encoding and decoding processes can be treated as a procedure for reconstructing observations. Hence, we can learn hidden representation $ \boldsymbol{h} $ by minimizing the errors between reconstructed observations $ \tilde{\boldsymbol{x}} $ and the origin inputs $ \boldsymbol{x} $ with the cost function

\begin{equation}
\theta^{\ast} = \arg \min_{\theta} E(\theta)
\end{equation} 
with $ E(\theta) = (\boldsymbol{x} - \tilde{\boldsymbol{x}})^{\top}(\boldsymbol{x} - \tilde{\boldsymbol{x}}) $ and $ \theta = \{\boldsymbol{w}_{(i,i+1)}, \boldsymbol{b}_{(i)}\}_{i=1}^{I} $ for all layers, and $ I $ is the number of layers. The discussion above indicates that researchers usually transform the representation learning process into a problem of training autoencoders, then select the output of the encoder $ \boldsymbol{h} $ (i.e., the layer of hidden presentation) in feature space $ \mathcal{H}\in \mathbb{R}^{r\times 1} $ for each driving encounter $ \boldsymbol{x} $ as the expected representations of this driving encounter, with $ r $ the dimension of the hidden layer.
%
%

\subsection{Distance-Based Measure}
Instead of treating the representation learning process as a black box, we use a mathematically rigorous way to capture their spatial relationship (e.g., approaching to each other fast or slowly) between two vehicles in a driving encounter. To begin with, we need to define a measure to gauge their geometrical distance. Generally, there are two straightforward ways to achieve this: dynamic time warping and normalized Euclidean distance.
%
%
%
%
\subsubsection{Dynamic Time Warping (DTW)}
Given a driving encounter observation $ \boldsymbol{x} $, the DTW\cite{muller2007dynamic} aims to measure the geometric relationship of the two-vehicle trajectories over time
\begin{equation*}
\begin{split}
\boldsymbol{p}^{(1)} & = (p^{(1)}_{1}, \cdots, p^{(1)}_{k},\cdots, p^{(1)}_{\bar{K}})\\
\boldsymbol{p}^{(2)} & = (p^{(2)}_{1}, \cdots, p^{(2)}_{k},\cdots, p^{(2)}_{\bar{K}})
\end{split}
\end{equation*}
where $ \bar{K} \in \mathbb{N} $ is the length of unified driving encounters and can be determined using training data. We define trajectory measure space $ \mathcal{P} \in \mathbb{R}^{\bar{K}\times \bar{K}} $, with $ p^{(1)}_{m}, p^{(2)}_{n}\in \mathcal{P} $ for $ m,n \in [1, \cdots, \bar{K}] $, where $n$ and $m$ are the positions of the two vehicles, respectively. The local distance of the positional point of one vehicle to one positional point of the other vehicle is defined and computed by a non-negative function $f$
%
%

\begin{equation}\label{eq:DTW_1}
f:\mathcal{P}\times \mathcal{P} \rightarrow \mathbb{R}_{\geq 0}
\end{equation} 
Typically, if $ p^{(1)}_{m} $ and $ p^{(2)}_{n} $ are close to each other, $ f (p^{(1)}_{m}, p^{(2)}_{n}) $ gets a small value, and otherwise $ f (p^{(1)}_{m}, p^{(2)}_{n}) $ attains a large value. Thus calculating the distance of pairwise elements in the two vehicle trajectories $ \boldsymbol{p}^{(1)} $ and $ \boldsymbol{p}^{(2)} $ allows to represent the geometry of driving encounter $ \boldsymbol{x} $ through a feature matrix $ \boldsymbol{F} \in \mathbb{R}^{\bar{K}\times \bar{K}} $ 

\begin{equation}\label{eq:DTW_2}
\boldsymbol{F}(m,n):=f(p^{(1)}_{m}, p^{(2)}_{n})
\end{equation}
with the distance measure function $f$. Here, the Euclidean distance is selected to compute the local distance, 

\begin{equation}\label{eq:DTW}
f(p^{(1)}_{m}, p^{(2)}_{n}) = \parallel p^{(1)}_{m}- p^{(2)}_{n}\parallel_{2}
\end{equation}
where $\|\cdot\|_{2} $ indicate the Euclidean distance between two points.

\subsubsection{Normalized Euclidean Distance (NED)}  The other efficient and straightforward way to measure the distance of temporal-pairwise positions of two vehicles at time $k$ is to apply an Euclidean distance directly. Thus, given a driving encounter, we can obtain a distance vector $\boldsymbol{f} = [f_{1}, f_{2}, \cdots, f_{\bar{K}}]$ to describe the relationship of the two vehicles, where

\begin{equation}\label{eq:NED}
f_{k} = \|p^{(1)}_{k} - p^{(2)}_{k}\|_{2}
\end{equation}
Then we normalized the distance vector as $\boldsymbol{F} = \frac{\boldsymbol{f}}{\max(\boldsymbol{f})}$. Therefore, we can get the feature $ \boldsymbol{F} $ of this driving encounter using the DTW and NED. 

In addition to DTW and NED, there exist other kinds of distance measures such as LCSS and Edit Distance for Real sequence (EDR)\cite{besse2016review}. They discretize the trajectories of driving encounter and take account of the number of occurrences, but the Euclidean distance between matched segments does not match a predefined spatial threshold. Compared to DTW, both of LCSS and EDR are sensitive to the spatial threshold\cite{besse2016review}, i.e., a large threshold value indicates a high acceptation of differences in trajectories and otherwise, low tolerance of differences. Therefore, we selected the DTW and NED instead of LCSS and EDR in this paper.

\subsection{Shape-Based Measure}
Different from the distance-based measure, the shape-based measure aims to capture the geometric features of two single trajectories\cite{besse2016review,besse2017destination}. The most well-known algorithms are the Hausdorff distance \cite{taha2015efficient}, the Fr\'{e}chet distance\cite{aronov2006frechet}, and the symmetrized segment-path distance (SSPD) \cite{besse2016review}. These methods have registered a high level of performance for describing the geometric features of a single trajectory; however, they are not suitable for capturing the relationship between driving encounters consisting of a pair of vehicle trajectories, since their output is a metric and easy to measure the shape-similarity level of individual trajectories but does not work for multi-vehicle trajectories. Even with the same metric value, the driving encounters could be significantly different from each other in terms of shape. Therefore, the shape-based measure is not used in this paper.

\subsection{Summary}
According to the above discussion, we select the deep neural network-based autoencoders and the distance-based measure to learn representations of driving encounters. Besides, the autoencoder can capture underlying information through dimension reduction and hence can potentially extract meaningful representations from the results of DTW or NED. There are five possible combinations of deep autoencoders and distance-based measure to learn representations of driving encounters, detailed as follows:
%
%
\begin{enumerate}
	\item DTW: Only using the output of DTW as representations,
	
	\begin{equation*}
	\mathrm{DTW}: \boldsymbol{x} \xrightarrow{\mathrm{Eq.} (\ref{eq:DTW_2})} \boldsymbol{F}
	\end{equation*}
	\item NED: Only using the output of NED as representations,
	
	\begin{equation*}
	\mathrm{NED}: \boldsymbol{x}\xrightarrow{\mathrm{Eq.}(\ref{eq:NED})} \boldsymbol{F}
	\end{equation*}
	\item LSTMAE: Applying LSTMAE to extract the representations of driving encounters, 
	
	\begin{equation*}
	\mathrm{LSTMAE}: \boldsymbol{x} \xrightarrow{\mathrm{LSTMAE}} \boldsymbol{h} 
	\end{equation*}
	\item DTW-ConvAE: Combing ConvAE with DTW to extract representations,
	
	\begin{equation*}
	\mathrm{DTW-ConvAE}: \boldsymbol{x}\xrightarrow{\mathrm{Eq.}(\ref{eq:DTW_2})}  \boldsymbol{F} \xrightarrow{\mathrm{ConvAE}} \boldsymbol{h} 
	\end{equation*}

	\item NED-LSTMAE: Combine LSTMAE with NED to extract representations,
	
	\begin{equation*}
	\mathrm{NED-LSTMAE}: \boldsymbol{x}\xrightarrow{\mathrm{Eq.}(\ref{eq:NED})}  \boldsymbol{F}\xrightarrow{\mathrm{LSTMAE}} \boldsymbol{h}
	\end{equation*}
\end{enumerate}
To understand this efficiently, we display and compare the input and output for each method of extracting representations in Table \ref{Table1}. For the ease of representation, we denoted $ \mathcal{F}_{\boldsymbol{x}_{i}} $ as the extracted representation $ \boldsymbol{F} $ or $ \boldsymbol{h} $ of driving encounter $ \boldsymbol{x}_{i} $ for all approaches in Table \ref{Table1}, where $K$ is the dimension of inputs and $r$ is the dimension of the learned representations.

\begin{table}[t]
	\centering
	\caption{Input and Output of Representation Learning Approaches}
	\label{Table1}
	\begin{tabular}{lll}
		\hline\hline
		Method & Input observations & Output features\\
		\hline
		DTW & $ \boldsymbol{x} \in \mathbb{R}^{\bar{K}\times d}$ & $ \boldsymbol{F} \in \mathbb{R}^{\bar{K}\times \bar{K}}$ \\
		NED & $ \boldsymbol{x} \in \mathbb{R}^{\bar{K}\times 1}$ & $ \boldsymbol{F} \in \mathbb{R}^{\bar{K}\times 1}$ \\
		LSTMAE & $ \boldsymbol{F} \in \mathbb{R}^{\bar{K}\times 1}$ & $ \boldsymbol{h}\in \mathbb{R}^{r\times 1} $ \\
		DTW-ConvAE & $ \boldsymbol{F} \in \mathbb{R}^{\bar{K}\times \bar{K}}$ & $ \boldsymbol{h}\in \mathbb{R}^{r\times 1} $ \\
		NED-LSTMAE & $ \boldsymbol{F} \in \mathbb{R}^{\bar{K}\times 1}$ & $ \boldsymbol{h}\in \mathbb{R}^{r\times 1} $ \\
		\hline\hline
	\end{tabular}
\end{table}

\section{Clustering}
This section shall introduce the clustering algorithms for the learned representations. We first detail the clustering method and then introduce the performance evaluation criteria of clustering results. Time-series trajectories clustering is very ubiquitous\cite{liao2005clustering}, and many clustering algorithms have been developed to solve related problems\cite{xu2005survey}. The learned representations of driving encounters restrict clustering method selection. Given the extracted representation of driving encounters in Table \ref{Table1}, we can classify them, i.e., gather associated driving encounters into groups. The learned representations are usually in a high dimension, which makes it practically intractable for specific clustering approaches such as DBSCAN (i.e., density-based spatial clustering of applications with noise).  In this paper, we prefer the  $ k $-means clustering ($ k $-MC)\cite{kanungo2002efficient} for simplicity and scalability. 
%
%

During the clustering procedure, we have limited prior knowledge about the number of groups. Therefore, we need a criterion to evaluate the clustering performance and consequently select an appropriate cluster number. We aim to gather driving encounters with similar characteristics into one group that is different from other groups. Hence, the quality of clustering results can be assessed by checking the between and within cluster variances of the obtained clusters. On the other hand, the variance of the elements in the same groups should be as small as possible. According to references \cite{xu2005survey,besse2016review,liao2005clustering}, we define the within-cluster (WC) variance and between-cluster (BC) variance, which requires the computation of a mean value of extracted features. Directly computing the mean of driving encounters is intractable since it is inconceivable and meaningless to calculate the mean of trajectories in driving encounters. Instead of using the multi-vehicle trajectories to evaluate the BC and WC variances directly, we computed the mean of the extracted representations (denoted as $ \bar{\mathcal{F}}_{\mathcal{X}} $) by averaging all extracted representations of driving encounters belong to cluster $ \mathcal{X} $. Let $ \mathcal{X}_{j}$ be the set of driving encounters in the $j$-th clusters. Assuming that we obtain $J$ clusters for $N$ driver encounters, then the BC variance and WC variance are computed by

\begin{subequations}
	\begin{align}
	BC & = \frac{1}{J-1} \sum_{j=1}^{J}\mathcal{D}(\bar{\mathcal{F}}_{\mathcal{X}}, \bar{\mathcal{F}}_{\mathcal{X}_{j}})\\
	WC & = \frac{1}{N-J} \sum_{j=1}^{J}\sum_{i}^{|\mathcal{X}_{j}|}\frac{1}{| \mathcal{X}_{j} |}\mathcal{D}(\mathcal{F}_{\boldsymbol{x}_{j,i}}, \bar{\mathcal{F}}_{\mathcal{X}_{j}})
	\end{align}
\end{subequations}
where $ \mathcal{D}(\cdot,\cdot) $ is the Euclidean distance measure, $ |\mathcal{X}_{j}| $ represents the number of driving encounters in the cluster $ \mathcal{X}_{j} $, $\mathcal{F}_{\boldsymbol{x}_{j,i}}$ is the feature representation of the $i$-th driving encounter in the $j$-th cluster, and  $\bar{\mathcal{F}}_{\mathcal{X}_{j}}$ is the mean of the feature representations of all driving encounters in the  $j$-th cluster. Our goal is to maximize the BC variance and minimize the WC variance; however, the BC and WC variances of representations extracted by different approaches are usually in different scales, which makes it meaningless to compare their BC and WC values directly. In order to make performance comparison tractable, we normalized the BC and WC variances by defining their relative metrics as follows:

\begin{equation}
\lambda_{BC} = \frac{BC}{WC+BC}
\end{equation}

\begin{equation}
\lambda_{WC} = \frac{WC}{WC+BC}
\end{equation}
In this way, the BC and WC variances between approaches become comparable by scaling their values between [0, 1] with respect to each approach. A large value of $ \lambda_{BC} $ indicates a large distance between clusters and otherwise, indicates a small distance between clusters. 

\section{Experiment and Data Collection}

\subsection{Data Collection and Analysis}
%
%
All the driving encounter data were collected from naturalistic settings supported by the University of Michigan Safety Pilot Model Development (SPMD) program\cite{wang2017much}. This SPMD database was collected by the University of Michigan Transportation Research Institute (UMTRI) and provided driving data logged in the last three years in Ann Arbor area, covering about 3,500 equipped vehicles and 6-million trips in total. The onboard GPS start to collect latitude and longitude information of each vehicle while igniting the equipped vehicle. The data were collected at a sampling frequency of 10 Hz. 

\begin{figure}[t]
	\centering
	\includegraphics[width = \linewidth]{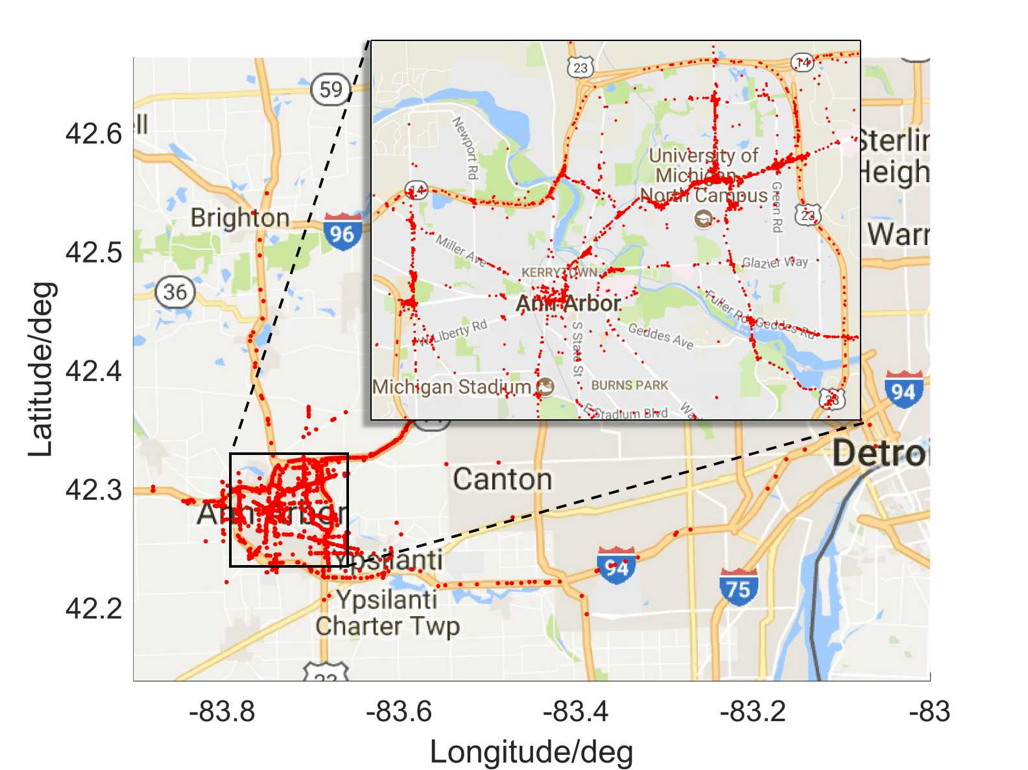}
	\caption{Scatter plot of driving encounters in Ann Arbor on the Google Map.}
	\label{fig:DEmap}
\end{figure}

\begin{figure}[t]
	\centering
	\includegraphics[width = \linewidth]{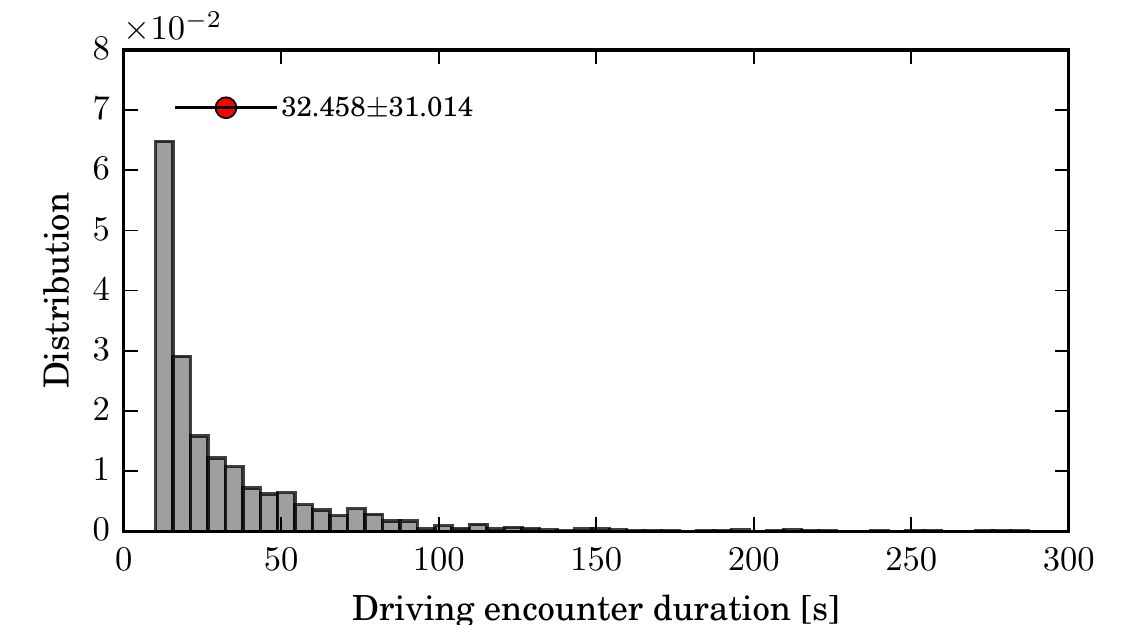}
	\caption{Distribution of origin driving encounter length in seconds.}
	\label{fig:drivingencounterdistribution}
\end{figure}

We searched the dataset of 100,000 trips, collected from 1900 vehicles with 12-day runs. The trajectory information we extracted includes latitude, longitude, the speed of the vehicles. The selection range was restricted to an urban area with the latitude in (-83.82, -83.64) and the longitude in (42.22, 42.34), as shown in Fig. \ref{fig:DEmap}. The vehicle encounter refers to as the scenario where two vehicles are close to each other, and the relative Euclidean distance between them is smaller than 100 m. The dots indicate the position of the vehicle at every sample time. After querying from the SPMD database, we got 49,998 such vehicle encounters. 
In the case where the distance between two vehicles is less than 100 m, then for a short second, they were out of the range required to communicate with each other, which results in very short trajectories. In this paper, we mainly focus on the trajectory length longer than 10 s, which is meaningful for analysis and applications. Finally, we obtained 2,568 driving encounters for experimental validation which fit these criteria. Fig. \ref{fig:drivingencounterdistribution} displays the distribution of time duration of all origin driving encounters. 

\subsection{Experiment Procedure and Settings}

\subsubsection{Autoencoders}

In order to make feature extraction efficiently, we used the re-sampling method (mentioned in Section II-B) through an interpolation process to unify the driving encounter into equal length $\bar{K}$. Fig. \ref{fig:drivingencounterdistribution} displays that the length of driving encounters was diverse, and the mean value of them is around 30 s. In order to facilitate the training procedure, we empirically unified each driving encounter to a fixed length of 200 sample points using linear interpolation. Here, we selected the unified length of 200 with the fact that a small number of unified samples will reduce model accuracy while a high number of unified samples will increase computational burden without significant model performance improvement. We set all autoencoders with a symmetric structure based on our experiences, and more specifically, 
%
%

\begin{figure*}[!t]
	\centering
	\includegraphics[width = \textwidth]{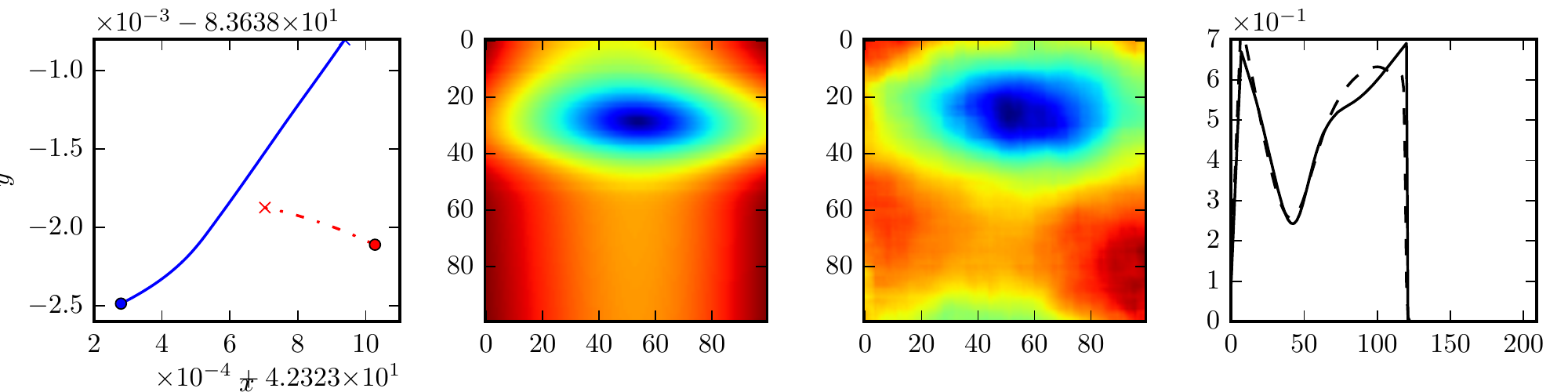}\\
	\includegraphics[width = \textwidth]{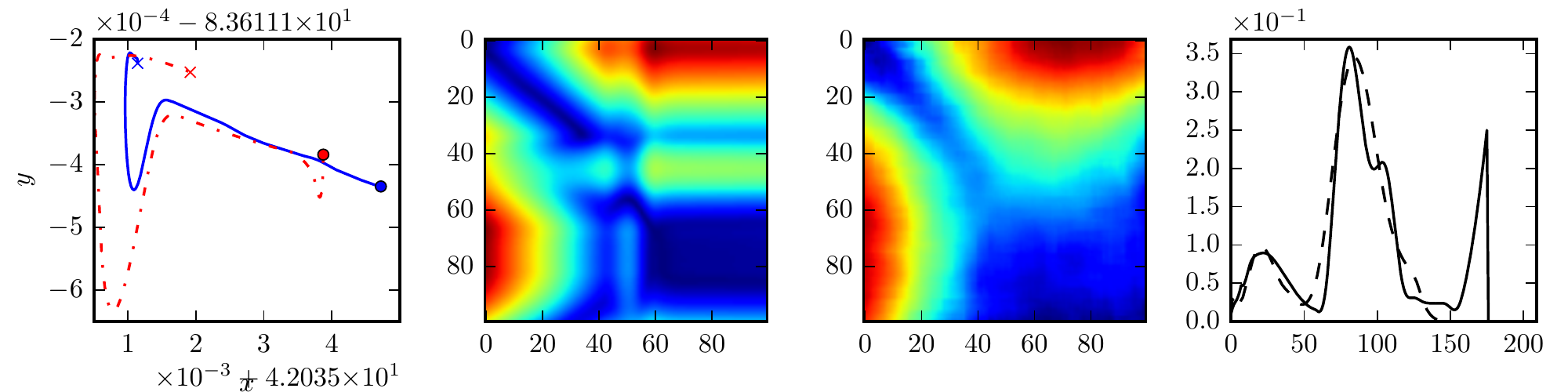}\\
	\includegraphics[width = \textwidth]{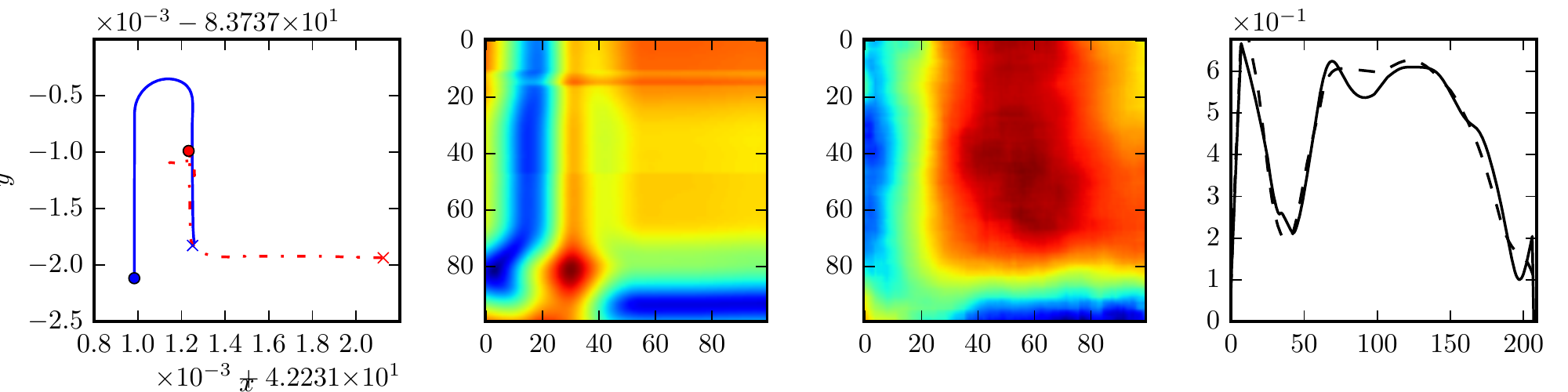}\\
	\includegraphics[width = \textwidth]{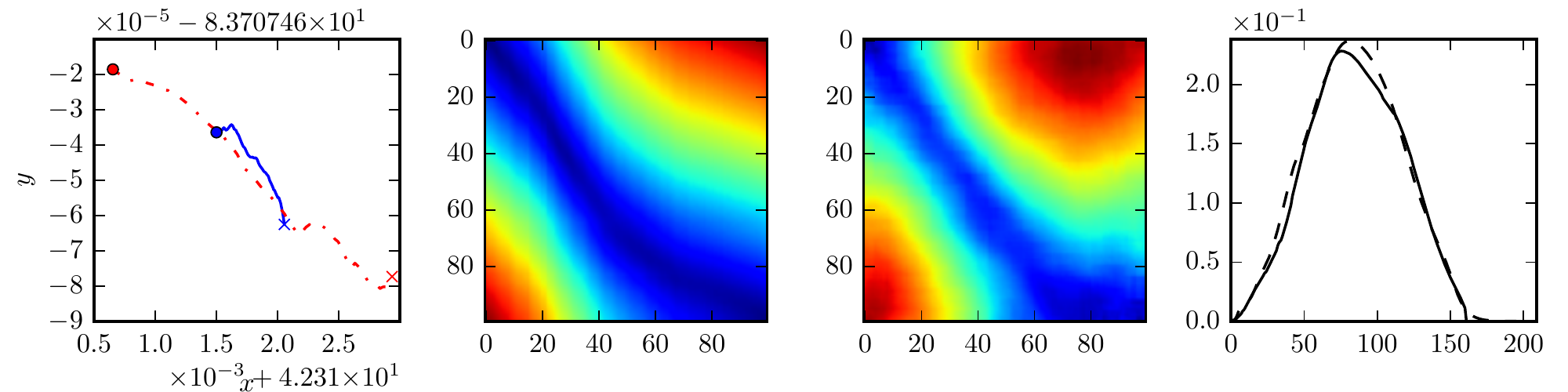}\\
	\caption{Examples of learning representations by different approaches. Left: Raw driving encounter GPS trajectories with start points (i.e., marked as dot) and end points (i.e., marked as cross). Mid-left: Extracted representation using DTW. Mid-right: Reconstructed representation of DTW using ConvAE. Right: Extracted representation using normalized Euclidean distance (NED) of two trajectories and its reconstructed representation from LSTMAE.}
	\label{fig:featureextractionresults}
\end{figure*}

\begin{itemize}
	\item LSTMAE: In this paper, we designed a five-layer ($ I=5 $) LSTMAE for learning the representations, where the third layer (i.e., the middle layer) is extracted as the hidden representation of driving encounters. The weights for each layer were initialized using random numbers between $1$ and $-1$ following a uniform distribution. The latent dimension (i.e., hidden layer) was set as 10, i.e., $ \boldsymbol{h}\in \mathbb{R}^{r\times 1}$ with $ r = 10 $.
	
	\item ConvAE: A five-layer ConvAE was designed to extract underlying features from outputs of DTW. The dimension of its hidden layer was also set as 10,  i.e., $ \boldsymbol{h}\in \mathbb{R}^{r\times 1}$ with $ r = 10 $. Applying the ConvAE to the high-dimension DTW feature matrix can significantly reduce the computation burden as the ConvAE can be typically used for dimensionality reduction.
\end{itemize}

After learning the hidden feature $ \mathcal{F}_{x_{i}} $ of driving encounter $ \boldsymbol{x}_{i} $ using autoencoders, we then applied the $ k $-means clustering ($ k $-MC) approach to these extracted hidden representations to cluster associated driving encounters. 
\subsubsection{DTW}
Similar to autoencoders, the length of driving encounters was unified to 100 for DTW. Thus we can compute the representation $ \mathcal{F}_{\boldsymbol{x}} \in \mathbb{R}^{100\times 100} $ using (\ref{eq:DTW}) for each driving encounter. Then we apply the clustering algorithms to all $\mathcal{F}_{\boldsymbol{x}}$.

\subsubsection{NED} For the NED approach, the representation can be directly computed through (\ref{eq:NED}) based on the unified driving encounters. Then, the output can be directly used for clustering and feeding into LSTMAE to obtain a low dimension representation (i.e., NED-LSTMAE).

For all learned representations, we do have limited prior knowledge of what values of $ k \in \mathbb{N}^{+}$ should be set. In order to determine $ k $, we applied clustering algorithms to different $ k $ and computed their performance metrics $ \lambda_{BC} $ and $ \lambda_{WC} $. 


\begin{figure}[t]
	\centering
	\subfloat[DTW-ConvAE]{\includegraphics[width = \linewidth]{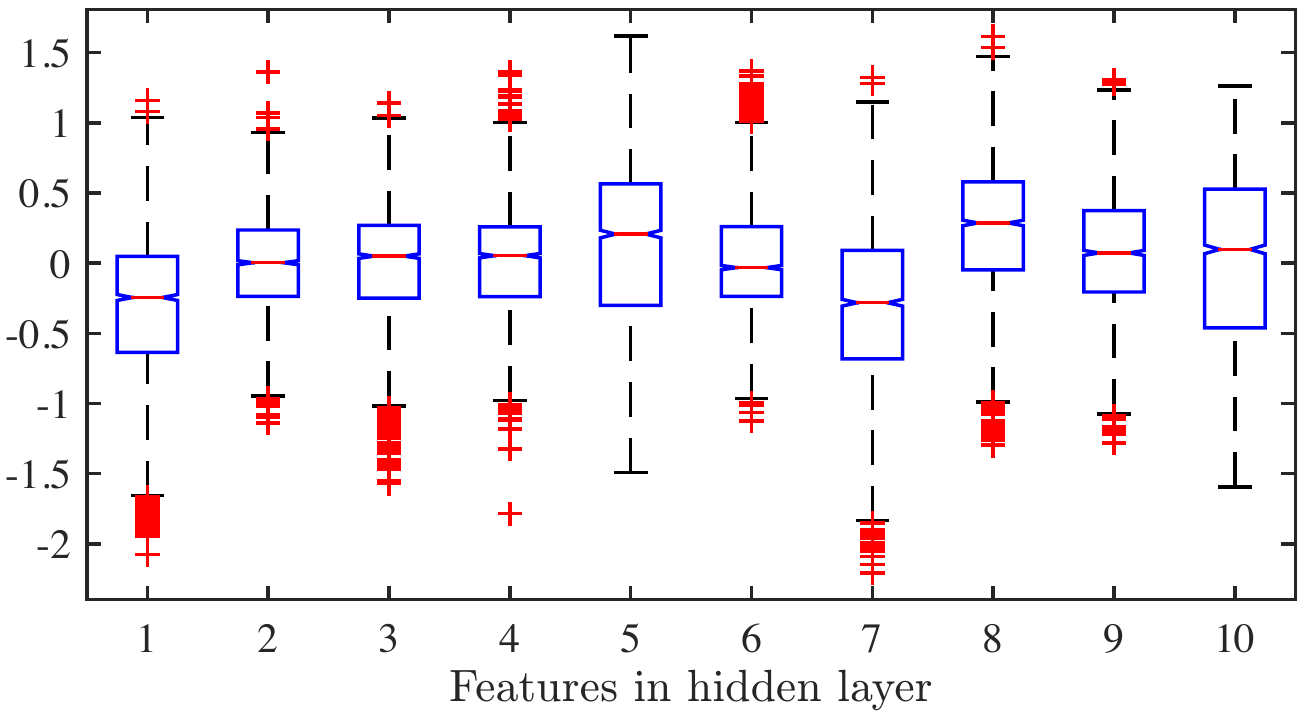}}
	\\
	\subfloat[NED-LSTMAE]{\includegraphics[width = \linewidth]{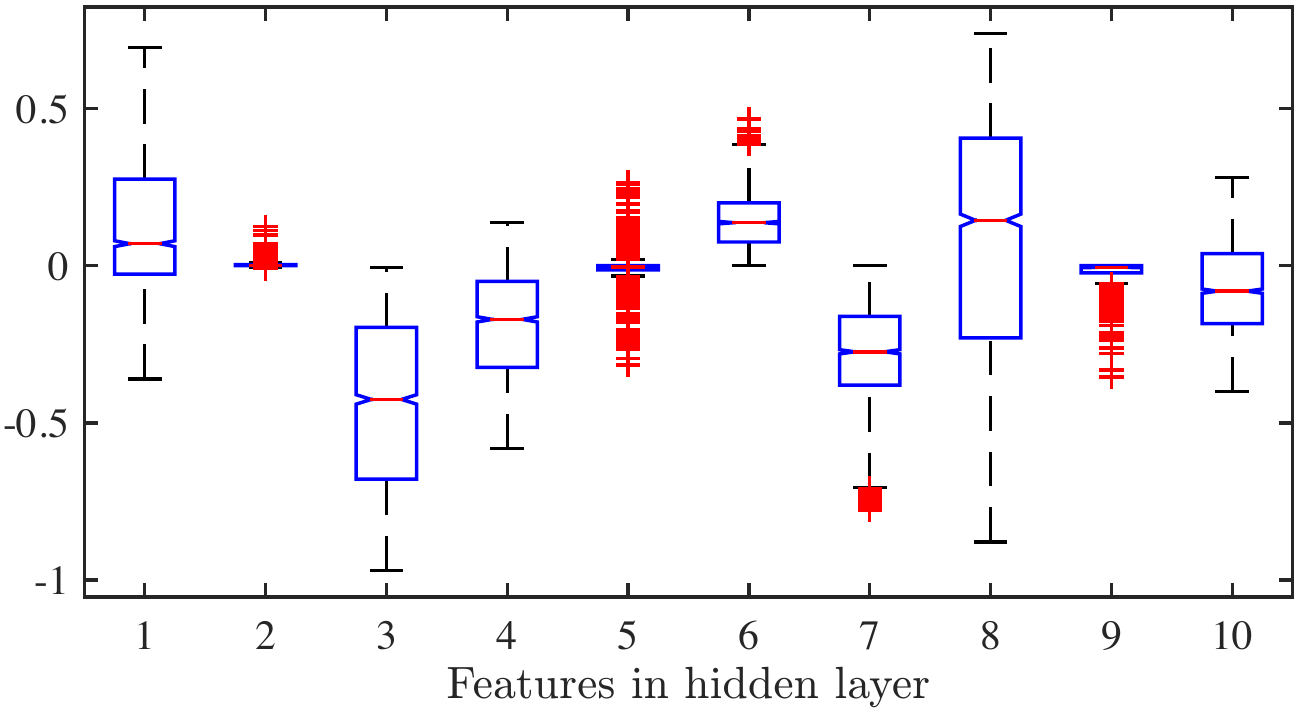}}
	\\
	\subfloat[LSTMAE]{\includegraphics[width = \linewidth]{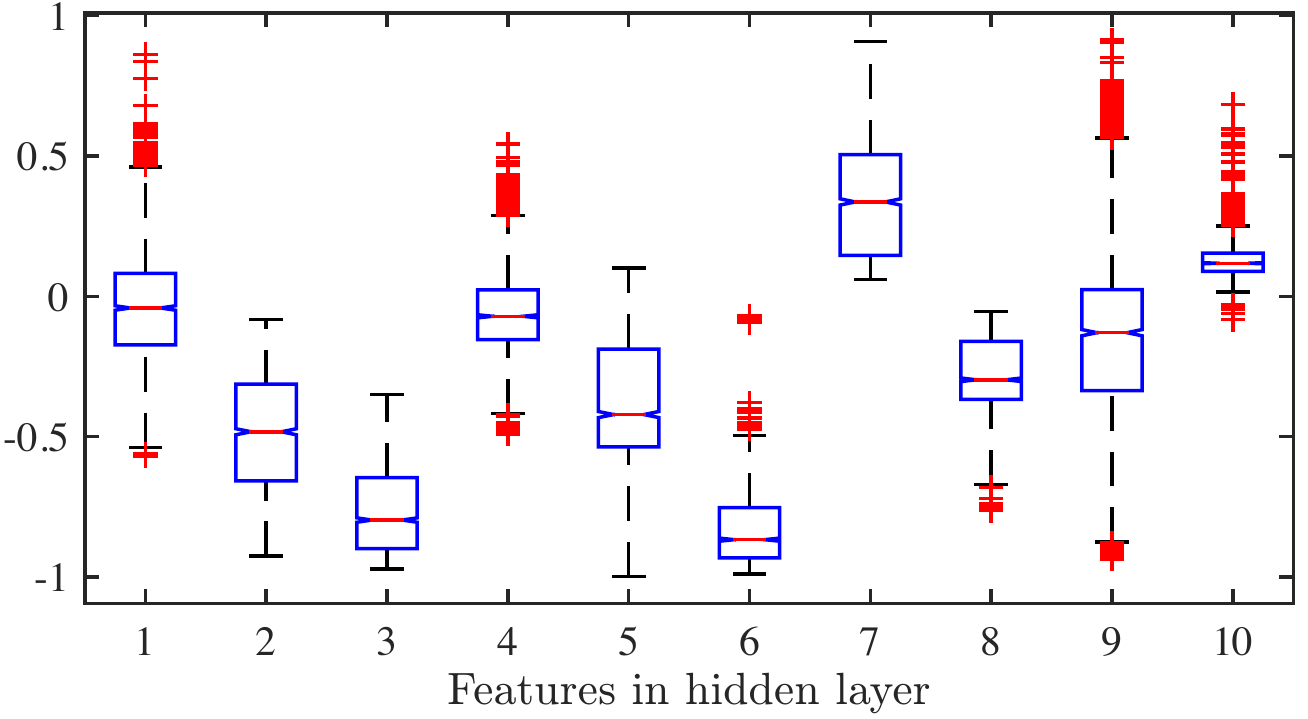}}
	\\
	\caption{Statistical results of extracted hidden features $ \boldsymbol{h} \in \mathbb{R}^{10\times 1} $ for 2,568 driving encounters using three autoencoders: (a) DTW-ConAE, (b) NED-LSTMAE, and (c) LSTMAE.}
	\label{fig:featuresautoencoders}
\end{figure}

\section{Result Discussion and Analysis}
This section will present and analyze the experiment results, including representation extraction results and clustering performance evaluation and comparison.
\subsection{Representation Learning and Analysis}


Fig. \ref{fig:featureextractionresults} displays four examples of learned representations using DTW and NED and the associated reconstructed results of using LSTMAE and ConvAE. The second column displays the DTW representations of associated driving encounter trajectories (the first column), and the third column displays the reconstruction results of using ConvAE from the hidden representation, $\boldsymbol{h}$. The second and third columns show similar feature matrix values (represented by color), indicating that the ConvAE can completely reconstruct the representations of DTW. On the other hand, the fourth column represents the NED results (solid line) and the reconstruction of NED outputs (dash line), and we can see that they match well with each other. In summary, the representations in the hidden layers of autoencoders (i.e., ConvAE and LSTMAE) can represent the associated driving encounters.

\begin{figure}[t]
	\centering
	\includegraphics[width = \linewidth]{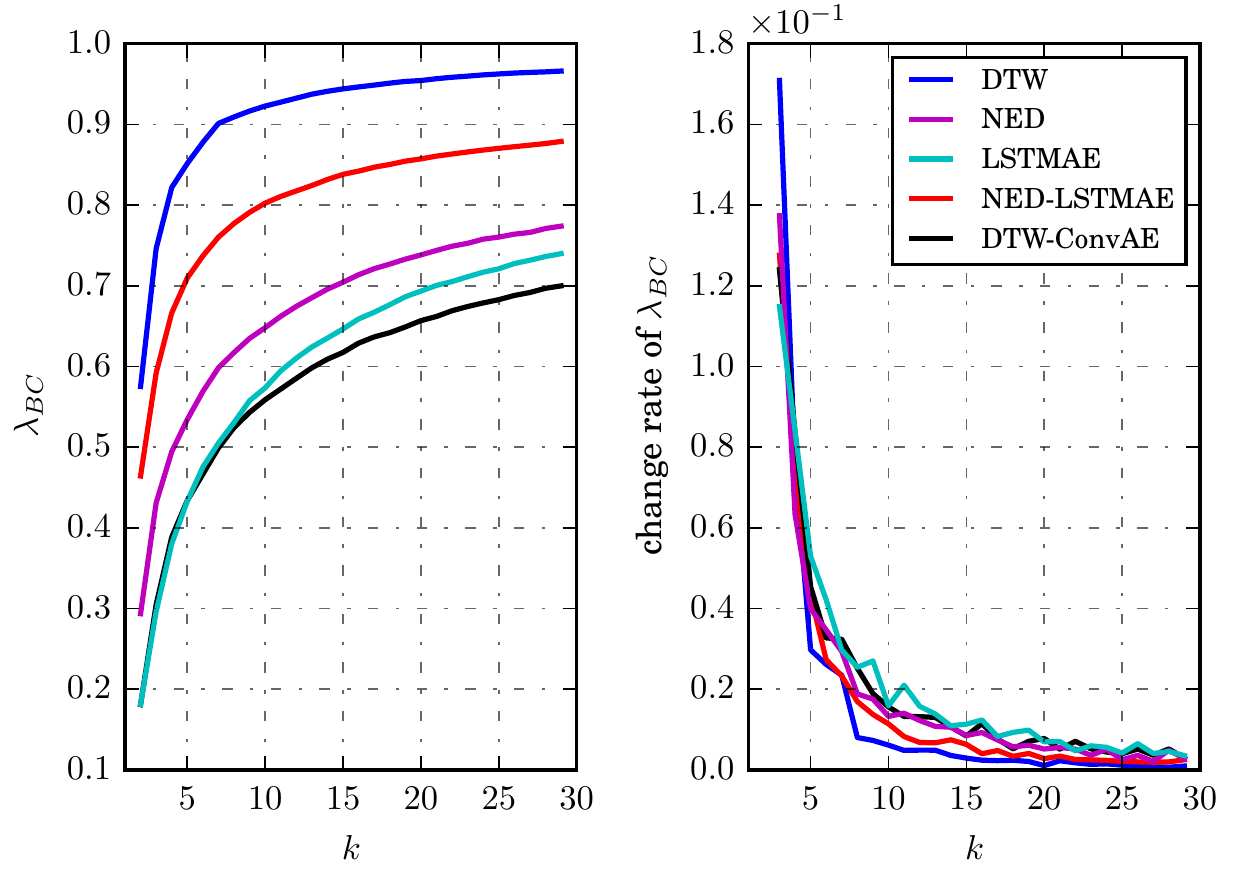}\\
	\includegraphics[width = \linewidth]{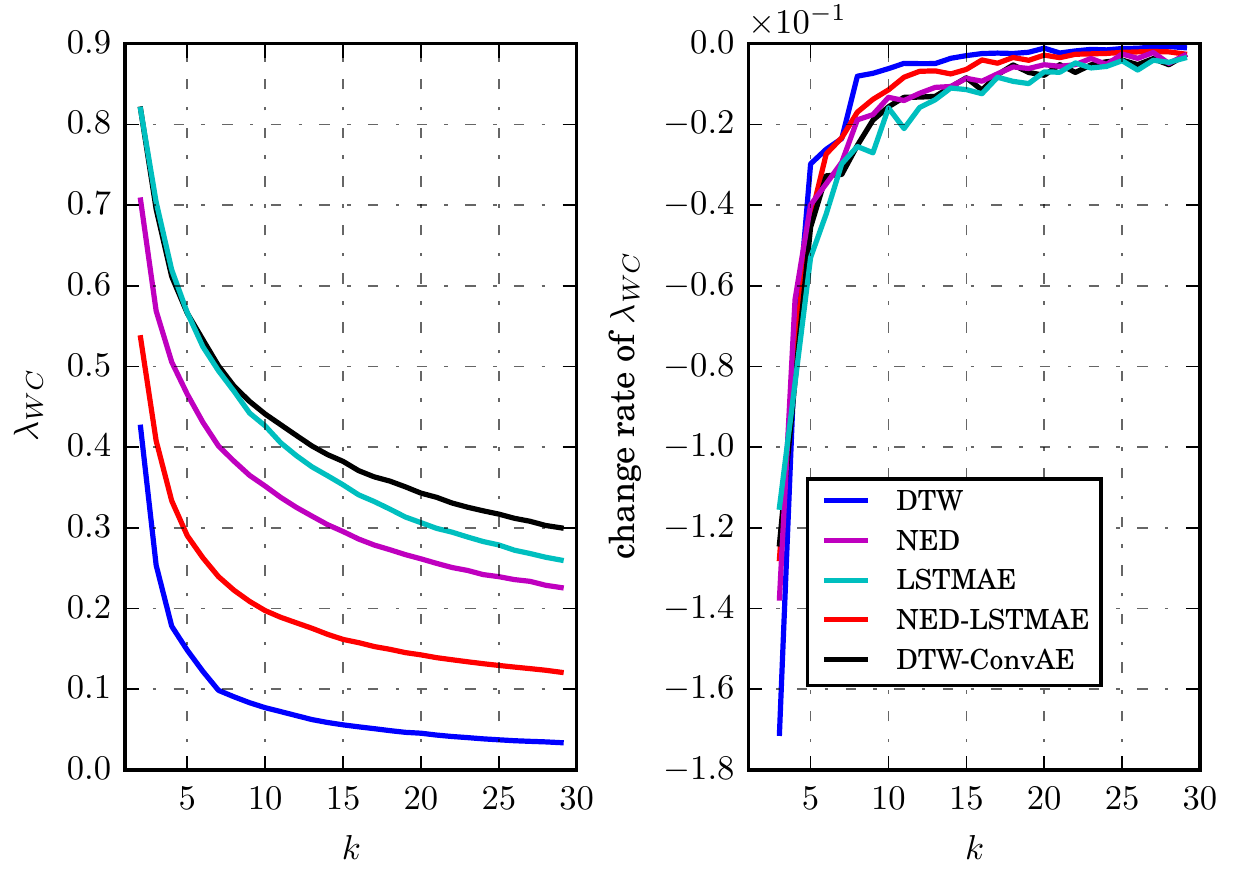}\\
	\caption{The performance of BC (top) and WC (bottom) distances using five representation extraction approaches (i.e., DTW, NED, LSTMAE, DTW-ConvAE, and NED-LSTMAE) with $ k $-MC.}
	\label{fig:BCWC}
\end{figure}

\subsubsection{Using DTW and NED}
Fig. \ref{fig:featureextractionresults} shows examples of the learned representations using DTW and NED. It can be known that both DTW and NED methods can capture the distance information of two vehicle trajectories in individual driving encounters. In the extracted representations of using DTW, deep red indicates a large distance of two vehicles and deep blue indicates a small distance. The DTW can capture the spatial information of all positions and the dynamic information over the whole trajectory since it computes the distance of trajectories over time, while the NED does not capture the dynamic information over temporal space. In addition, Fig. \ref{fig:featureextractionresults} presents the associated reconstructed outputs of DTW and NED. Experimental results demonstrate that the ConvAE and LSTMAE can capture the underlying information of the extracted representations using DTW and NED, respectively.

\begin{figure*}[t]
	\centering
	\includegraphics[width = 0.85\linewidth]{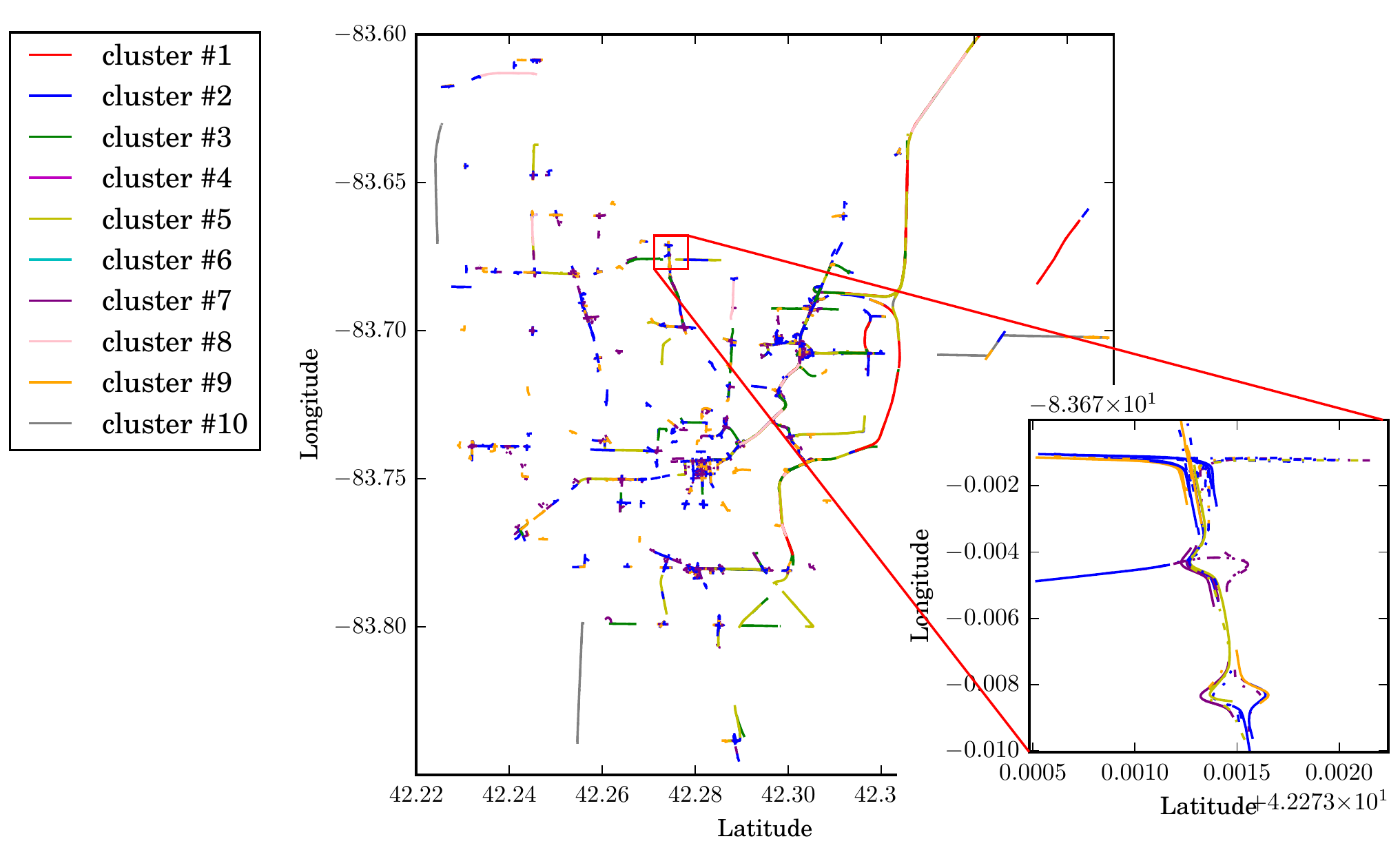}
	\caption{GPS data visualization of the clustering results using DTW-$ k $MC with $ k=10 $.}
	\label{fig:clustermap}
\end{figure*}

\subsubsection{Using Autoencoders}
Each driving encounter results in a ten-dimensional representation vector $\boldsymbol{h}\in \mathbb{R}^{10\times 1}$ from the designed autoencoders (i.e., ConvAE and LSTMAE). Fig. \ref{fig:featuresautoencoders} shows box plots of hidden layers in different autoencoders for all driving encounters. For each box, the central mark indicates the median, and the bottom and top edges of the box indicate the 25th and 75th percentiles, respectively. We found that DTW-ConvAE obtains less recognizable hidden representations; that is, distributions between each element of representation $ \boldsymbol{h} $ do not have significant differences, compared to NED-LSTMAE and LSTMAE. All elements in extracted representations using DTW-ConvAE have similar median values almost around zero and similar ranges of [25th, 75th] percentiles. For the NED-LSTMAE and LSTMAE methods, both of their hidden representations are recognizable, with median values and ranges of [25th, 75th] percentiles are sparsely distinctive. 

\subsection{Clustering Results and Analysis}

Based on the extracted representations, Fig. \ref{fig:BCWC} compares all approaches to cluster driving encounters by showing the scaled within-cluster and between-cluster metrics ($ \lambda_{BC} $ and $ \lambda_{WC} $) and their change rates over the number of clusters $ k $. We found that increasing the number of clusters would decrease the within-cluster distance while increasing the between-cluster distance. According to their change rate of $ \lambda_{BC} $ or $ \lambda_{WC} $, when the number of clusters is close to $ 10 $, their performance metrics would converge, implying that $ k = 10$ is the preferred selection. As we discussed before, the goal of clustering is to put homogeneous driving encounters into groups while maximizing the between-cluster distance and minimizing the within-cluster distance. Fig. \ref{fig:BCWC} indicates that the DTW-$ k $MC outperforms other counterparts, with $ \lambda_{BC} =  0.923$ at $ k = 10 $.

\begin{figure}[t]
	\centering
	\includegraphics[width = \linewidth]{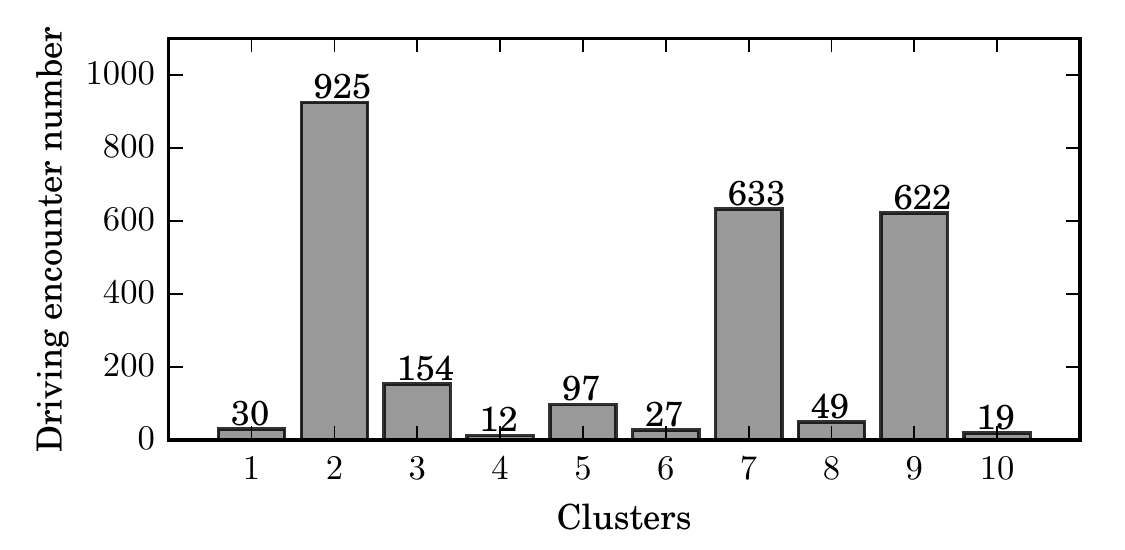}
	\caption{Clustering results of using DTW-$ k $MC with $ k=10 $.}
	\label{fig:clusters}
\end{figure}

For all autoencoders at $ k=10 $, the NED-LSTMAE-$ k $MC obtains the best performance with $ \lambda_{BC} =  0.803$, compared to LSTMAE-$ k $MC with $ \lambda_{BC} =0.574 $ and DTW-ConvAE-$ k $MC with $ \lambda_{BC} = 0.559 $. This can be explained according to Fig. \ref{fig:featuresautoencoders}: The DTW-ConvAE-$ k $MC (top in Fig. \ref{fig:featuresautoencoders}) obtains the worst performance as its extracted representations of driving encounters are not such recognizable and distinctive, compared with the other two approaches. For the distance-based measure, the DTW-$ k $MC outperforms the NED-$ k $MC with $ \lambda_{BC} = 0.648$, because the DTW can capture the dynamic features over time and the distance features of two trajectories, but the NED cannot.

Fig. \ref{fig:clustermap} visualizes the results of all clustered driving encounters with their GPS data. We only show the results with the optimal number of clusters $ k=10 $ because we have seen in Fig. \ref{fig:BCWC} that a plateau can be observed on the change rate of within-cluster criteria $ \lambda_{BC} $ with respect to the number of cluster starting at cluster sizes of 10 for all approaches. Fig. \ref{fig:clusters} lists the amount of driving encounters in each cluster. It can be found that some clusters covering very common driving encounter behaviors (e.g., cluster \#2, cluster \#7 and cluster \#9) and some clusters representing rare driving encounter behavior (e.g., cluster \#4 and cluster\#10) can be detected. For cluster \#1 (red) and cluster \#10 (gray), they represent two typical driving encounter scenarios, i.e. car-following on highways and driving with the opposite direction on highways, respectively. For the most occurred driving encounter scenario (cluster \#2), most of them are recorded at intersections, with a short duration, compared to cluster \#1 and cluster \#10. 

\subsection{Computing Efficiency Analysis}

\begin{table}[t]
	\centering
	\caption{Efficiency Comparison of Representation Extraction Approaches}
	\label{Table:computationefficiency}
	\begin{tabular}{ccc}
		\hline\hline
		Methods & Computing time & Unit \\
		\hline
        NED & 101.41 & second\\
		DTW & 120.77 & second\\
		LSTMAE & $\approx$ 10 & hour\\
		ConvAE & $\approx$ 5 & hour\\
		\hline\hline
	\end{tabular}
\end{table}

In order to evaluate the conservativeness of the algorithm and its applicability to real-time applications, its computational costs when implemented on a standard laptop computer are presented, and a comparison with other counterparts is given. All autoencoders were built using Python in Keras\footnote[1]{\url{https://blog.keras.io/building-autoencoders-in-keras.html}} with a tensorflow backend and all distance-based representation extraction algorithms were also programmed using Python. Table \ref{Table:computationefficiency} presents the average computation time for extracting representations on a standard laptop computer with an Intel Core i7 running at 2.5 GHz, with 16 GB of RAM. It can be seen that the NED and DTW algorithms can extract feature much faster than two others, only with 120 s for 2568 driving encounters. However, the autoencoders with neural nets require a few hours to extract representations. 

\subsection{Further Discussion}
This paper presents a comprehensive investigation of driving encounter classification through five approaches, which consists of two types: deep learning-based and distance-based. The distance-based method outperforms all other counterparts. However, some challenges still exist in our developed approaches and discussed as follows.

\subsubsection{Layers of Autoencoders}
For the deep learning-based approaches, we empirically set their hyperparameters, like the number of layers of decoder and encoder and the number of nodes in each layer. The hyperparameter selection in our paper mainly concerns two aspects: computational cost and model performance. Autoencoder with large numbers of layers could enhance model performance but at a significantly high computational cost. Therefore, in this paper, we selected a moderate number of layers to learn a hidden representation of driving encounters.

\subsubsection{Contextual Road Information}
This paper mainly utilized the multi-vehicle trajectories (GPS signals) without utilizing other information since GPS data is easy to obtain at a low cost such as via mobile-phone and equipped localization sensors. Our experiment validation did not consider other information such as contextual road information and vehicle speed, but it might extend our developed approach to other driving cases of including high-dimensional time-series data. For instance, if more road context information could be obtained such as highway and intersections (T-shape, Y-shape, and cross-shape, etc.), our developed approaches can help get insights into the diversity of driving encounters. Therefore, considering contextual road information will be one of our future work. Moreover, the feature selection can influence the clustering results, namely, adding more feature information (e.g., vehicle speed, contextual traffic, the number of agents in driving encounters, etc.) as inputs could change the final number of clusters.

\section{Conclusion}
This paper proposed a generic two-layer framework of clustering naturalistic driving encounters that consists of multi-vehicle trajectories. Two kinds of representation learning -- deep autoencoders and distance-based were developed and evaluated. Experimental results demonstrate that our proposed framework can gather homogeneous driving encounters into groups considering the spatiotemporal relationship between vehicles. Besides, the dynamic time warping (DTW) approach with $ k $-means clustering method outperforms other counterparts, in terms of both clustering performance and computational burden. We finally evaluated the performance of each proposed approach and confirmed that when we are only concerned with the vehicle trajectories, an acceptable and preferable cluster number is 10 in our database. These clustering results could provide insights into interactive driver behaviors in different scenarios and help to design a friendly and efficient decision-making policy for intelligent vehicles.

\section*{Acknowledgment}
Toyota Research Institute (``TRI") provided funds to assist the authors with their research but this article solely reflects the opinions and conclusions of its authors and not TRI or any other Toyota entity.

\ifCLASSOPTIONcaptionsoff
  \newpage
\fi



\bibliographystyle{IEEEtran}
\bibliography{ref.bib}
%
%
%

%

\begin{IEEEbiography}[{\includegraphics[width=1in,height=1.25in,clip,keepaspectratio]{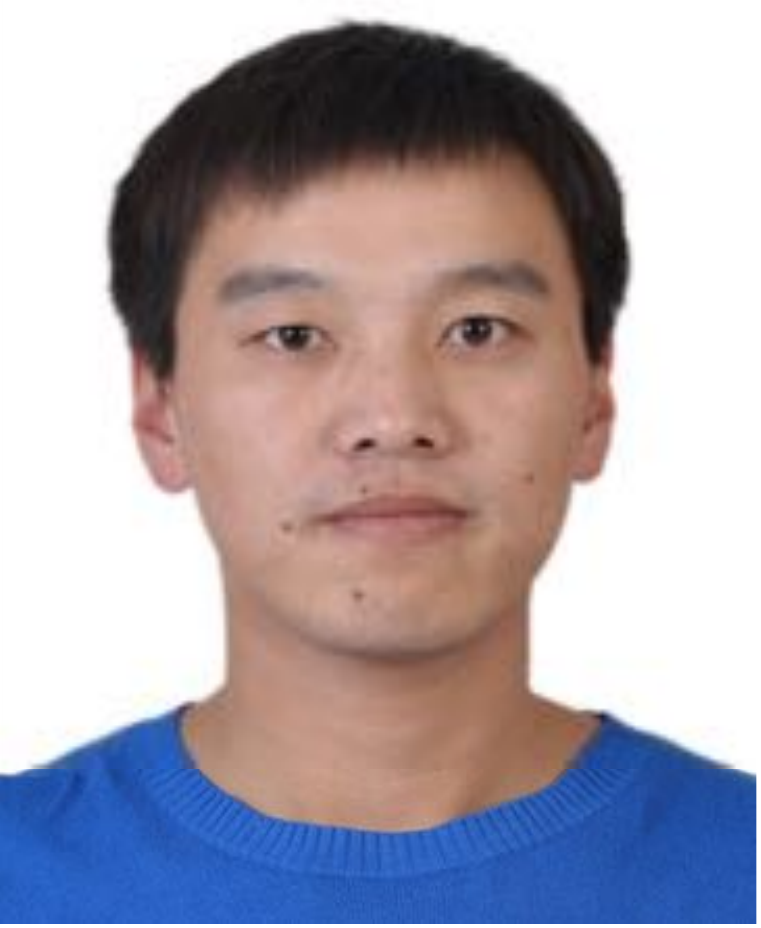}}]{Wenshuo Wang} (S'15-M'18) received his Ph.D.
degree in mechanical engineering from the Beijing
Institute of Technology (BIT) in June, 2018. He is
now working as a postdoc at Carnegie Mellon University
(CMU), Pittsburgh, PA. He also worked as a
Research Scholar in the Department of Mechanical
Engineering, University of California at Berkeley
(UCB) from September 2015 to September 2017
and in the Department of Mechanical Engineering,
University of Michigan (UM), Ann Arbor, from
September 2017 to July 2018. His research interests
include nonparametric Bayesian learning, driver models, human-vehicle interactions,
and the recognition and application of human driving characteristics.
\end{IEEEbiography}

\begin{IEEEbiography}[{\includegraphics[width=1in,height=1.25in,clip,keepaspectratio]{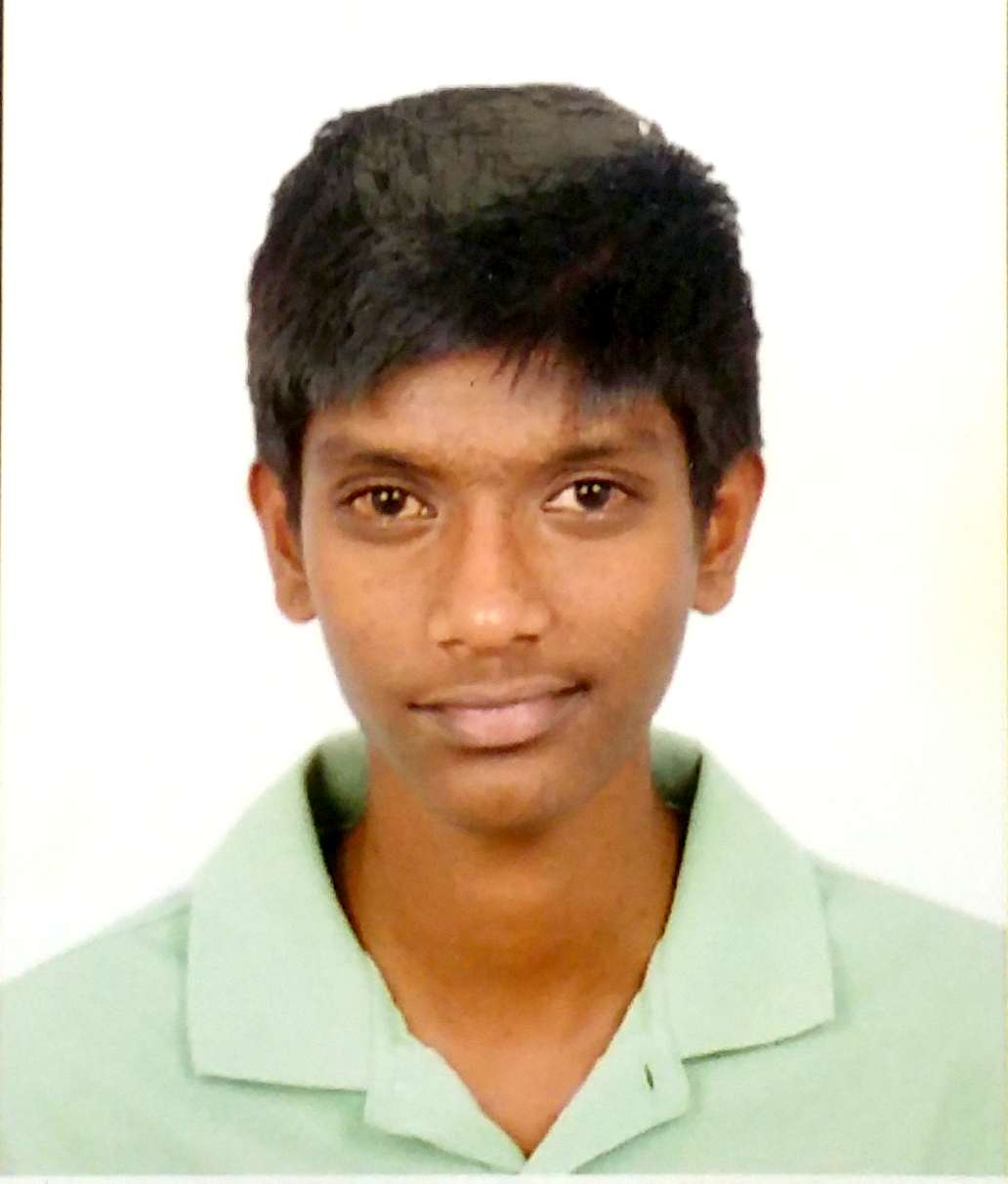}}]{Aditya Ramesh} 
is a Master student with the Department of Electrical Engineering and Compuer Science (EECS), University of Michigan, Ann Arbor, MI, USA. 

His research interests include unsupervised learning and autonomous driving.
\end{IEEEbiography}


\begin{IEEEbiography}[{\includegraphics[width=1in,height=1.25in,clip,keepaspectratio]{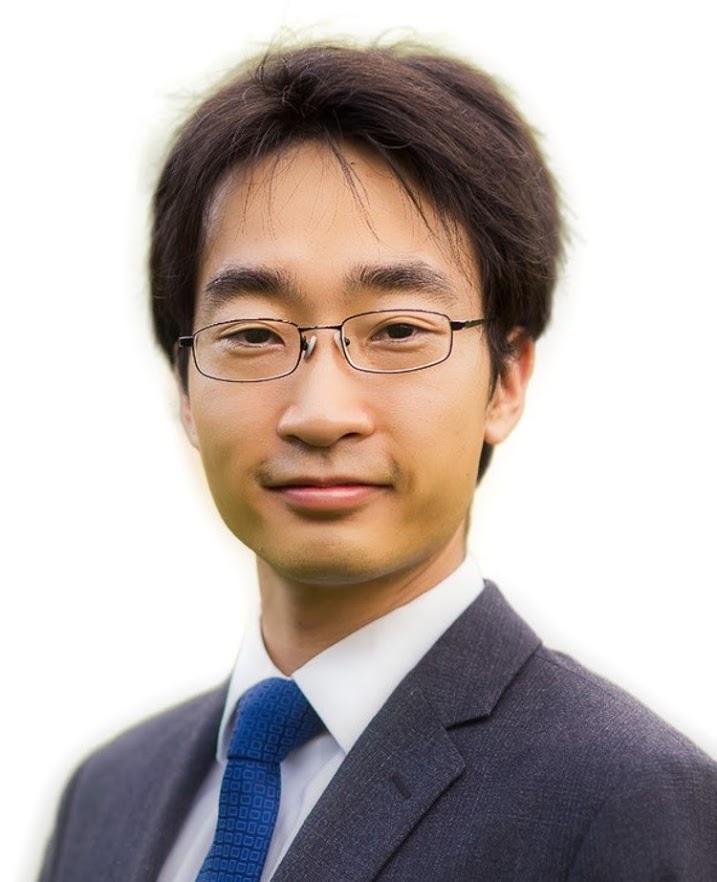}}]{Ding Zhao} received his Ph.D. degree in 2016 from
the University of Michigan, Ann Arbor. He is currently an Assistant Professor at Department of Mechanical Engineering, Carnegie Mellon University. His research focuses on the intersection of robotics, machine learning, and design, with applications on autonomous driving, connected/smart city, energy efficiency, human-machine interaction, cybersecurity, and big data analytics.
\end{IEEEbiography}

%
%




\end{document}

%% file: figures/drivingencounter.tex
\tikzset{
	on each segment/.style={
		decorate,
		decoration={
			show path construction,
			moveto code={},
			lineto code={
				\path [#1]
				(\tikzinputsegmentfirst) -- (\tikzinputsegmentlast);
			},
			curveto code={
				\path [#1] (\tikzinputsegmentfirst)
				.. controls
				(\tikzinputsegmentsupporta) and (\tikzinputsegmentsupportb)
				..
				(\tikzinputsegmentlast);
			},
			closepath code={
				\path [#1]
				(\tikzinputsegmentfirst) -- (\tikzinputsegmentlast);
			},
		},
	},
	mid arrow/.style={postaction={decorate,decoration={
				markings,
				mark=at position .5 with {\arrow[#1]{stealth}}
			}}},
		}

\begin{tikzpicture}
\draw [blue,dotted,postaction={on each segment={mid arrow=blue}}] (-3,1) to [out=10,in=170] (0,1);
\draw [blue, postaction={on each segment={mid arrow=blue}}] (0,1) to [out=0,in=170] (3,2);
\draw [blue, dotted, postaction={on each segment={mid arrow=blue}}] (3,2) to [out=-10,in=180] (4,2);
\fill [blue] (0,1) circle (0.05);
\fill [blue] (3,2) circle (0.05);
\node[text width=1cm, blue] at (0.2,0.5) {$ p_{k}^{(1)} $};
\node[text width=1cm, blue] at (3.2,1.5) {$ p_{k+1}^{(1)} $};
\node[inner sep=0pt, rotate = 5]at(-2.3,1.1){\includegraphics[width = 0.1\linewidth]{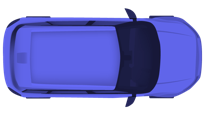}};
\node[text width = 2cm, blue] at (-2.,0.7) {\footnotesize Vehicle \#1};
\draw [red, dotted, postaction={on each segment={mid arrow=red}}] (-2.5,2) to [out=20, in = 200] (0.5, 1.5);
\draw [red, postaction={on each segment={mid arrow=red}}] (0.5,1.5) to [out=20, in = 200] (2.5, 2.5);
\draw [red, dotted, postaction={on each segment={mid arrow=red}}] (2.5,2.5) to [out=20, in = -200] (4, 2.5);
\fill [red] (0.5,1.5) circle (0.05);
\fill [red] (2.5,2.5) circle (0.05);
\node[text width = 1cm, red] at (0.7,2) {$ p^{(2)}_{k} $};
\node[text width = 1cm, red] at (2.7,3) {$ p^{(2)}_{k+1} $};
\node[text width = 2cm, red] at (-1.5,2.5) {\footnotesize Vehicle \#2};
\node[inner sep=0pt, rotate = -10]at(-1.8,2.){\includegraphics[width = 0.1\linewidth]{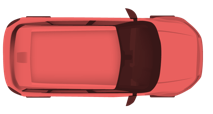}};
\end{tikzpicture}

%% file: bare_jrnl.bbl
\begin{thebibliography}{10}
\providecommand{\url}[1]{#1}
\csname url@samestyle\endcsname
\providecommand{\newblock}{\relax}
\providecommand{\bibinfo}[2]{#2}
\providecommand{\BIBentrySTDinterwordspacing}{\spaceskip=0pt\relax}
\providecommand{\BIBentryALTinterwordstretchfactor}{4}
\providecommand{\BIBentryALTinterwordspacing}{\spaceskip=\fontdimen2\font plus
\BIBentryALTinterwordstretchfactor\fontdimen3\font minus
  \fontdimen4\font\relax}
\providecommand{\BIBforeignlanguage}[2]{{%
\expandafter\ifx\csname l@#1\endcsname\relax
\typeout{** WARNING: IEEEtran.bst: No hyphenation pattern has been}%
\typeout{** loaded for the language `#1'. Using the pattern for}%
\typeout{** the default language instead.}%
\else
\language=\csname l@#1\endcsname
\fi
#2}}
\providecommand{\BIBdecl}{\relax}
\BIBdecl

\bibitem{deo2018would}
N.~Deo, A.~Rangesh, and M.~M. Trivedi, ``How would surround vehicles move? {A}
  unified framework for maneuver classification and motion prediction,''
  \emph{arXiv preprint arXiv:1801.06523}, 2018.

\bibitem{taylor2015method}
J.~Taylor, X.~Zhou, N.~M. Rouphail, and R.~J. Porter, ``Method for
  investigating intradriver heterogeneity using vehicle trajectory data: A
  dynamic time warping approach,'' \emph{Transportation Research Part B:
  Methodological}, vol.~73, pp. 59--80, 2015.

\bibitem{wang2017driving}
W.~Wang, J.~Xi, A.~Chong, and L.~Li, ``Driving style classification using a
  semisupervised support vector machine,'' \emph{IEEE Transactions on
  Human-Machine Systems}, vol.~47, no.~5, pp. 650--660, 2017.

\bibitem{besse2017destination}
P.~C. Besse, B.~Guillouet, J.-M. Loubes, and F.~Royer, ``Destination prediction
  by trajectory distribution-based model,'' \emph{IEEE Transactions on
  Intelligent Transportation Systems}, 2017.

\bibitem{piciarelli2008trajectory}
C.~Piciarelli, C.~Micheloni, and G.~L. Foresti, ``Trajectory-based anomalous
  event detection,'' \emph{IEEE Transactions on Circuits and Systems for video
  Technology}, vol.~18, no.~11, pp. 1544--1554, 2008.

\bibitem{sun2015trajectory}
Z.~Sun, P.~Hao, X.~J. Ban, and D.~Yang, ``Trajectory-based vehicle
  energy/emissions estimation for signalized arterials using mobile sensing
  data,'' \emph{Transportation Research Part D: Transport and Environment},
  vol.~34, pp. 27--40, 2015.

\bibitem{chen2017vehicle}
Z.-J. Chen, C.-Z. Wu, Y.-S. Zhang, Z.~Huang, J.-F. Jiang, N.-C. Lyu, and
  B.~Ran, ``Vehicle behavior learning via sparse reconstruction with $ \ell_2 -
  \ell_{p}$ minimization and trajectory similarity,'' \emph{IEEE Transactions
  on Intelligent Transportation Systems}, vol.~18, no.~2, pp. 236--247, 2017.

\bibitem{appenzeller2017scientists}
T.~Appenzeller, ``The scientists' apprentice,'' \emph{Science}, vol. 357, no.
  6346, pp. 16--17, 2017.

\bibitem{yuan2017review}
G.~Yuan, P.~Sun, J.~Zhao, D.~Li, and C.~Wang, ``A review of moving object
  trajectory clustering algorithms,'' \emph{Artificial Intelligence Review},
  vol.~47, no.~1, pp. 123--144, 2017.

\bibitem{feng2016survey}
Z.~Feng and Y.~Zhu, ``A survey on trajectory data mining: Techniques and
  applications,'' \emph{IEEE Access}, vol.~4, pp. 2056--2067, 2016.

\bibitem{besse2016review}
P.~C. Besse, B.~Guillouet, J.-M. Loubes, and F.~Royer, ``Review and perspective
  for distance-based clustering of vehicle trajectories,'' \emph{IEEE
  Transactions on Intelligent Transportation Systems}, vol.~17, no.~11, pp.
  3306--3317, 2016.

\bibitem{yao2018learning}
D.~Yao, C.~Zhang, Z.~Zhu, Q.~Hu, Z.~Wang, J.~Huang, and J.~Bi, ``Learning deep
  representation for trajectory clustering,'' \emph{Expert Systems}, DOI:
  10.1111/exsy.12252.

\bibitem{zhao2017road}
H.~Zhao, C.~Wang, Y.~Lin, F.~Guillemard, S.~Geronimi, and F.~Aioun, ``On-road
  vehicle trajectory collection and scene-based lane change analysis: Part
  {I},'' \emph{IEEE Transactions on Intelligent Transportation Systems},
  vol.~18, no.~1, pp. 192--205, 2017.

\bibitem{yao2017road}
W.~Yao, Q.~Zeng, Y.~Lin, D.~Xu, H.~Zhao, F.~Guillemard, S.~Geronimi, and
  F.~Aioun, ``On-road vehicle trajectory collection and scene-based lane change
  analysis: Part {II},'' \emph{IEEE Transactions on Intelligent Transportation
  Systems}, vol.~18, no.~1, pp. 206--220, 2017.

\bibitem{choong2017modeling}
M.~Y. Choong, L.~Angeline, R.~K.~Y. Chin, K.~B. Yeo, and K.~T.~K. Teo,
  ``Modeling of vehicle trajectory clustering based on lcss for traffic pattern
  extraction,'' in \emph{Automatic Control and Intelligent Systems (I2CACIS),
  2017 IEEE 2nd International Conference on}.\hskip 1em plus 0.5em minus
  0.4em\relax IEEE, 2017, pp. 74--79.

\bibitem{zhao2017trajectory}
P.~Zhao, K.~Qin, X.~Ye, Y.~Wang, and Y.~Chen, ``A trajectory clustering
  approach based on decision graph and data field for detecting hotspots,''
  \emph{International Journal of Geographical Information Science}, vol.~31,
  no.~6, pp. 1101--1127, 2017.

\bibitem{wang2017automatic}
J.~Wang, C.~Wang, X.~Song, and V.~Raghavan, ``Automatic intersection and
  traffic rule detection by mining motor-vehicle gps trajectories,''
  \emph{Computers, Environment and Urban Systems}, vol.~64, pp. 19--29, 2017.

\bibitem{zhan2017citywide}
X.~Zhan, Y.~Zheng, X.~Yi, and S.~V. Ukkusuri, ``Citywide traffic volume
  estimation using trajectory data,'' \emph{IEEE Transactions on Knowledge and
  Data Engineering}, vol.~29, no.~2, pp. 272--285, 2017.

\bibitem{kim2015spatial}
J.~Kim and H.~S. Mahmassani, ``Spatial and temporal characterization of travel
  patterns in a traffic network using vehicle trajectories,''
  \emph{Transportation Research Part C: Emerging Technologies}, vol.~59, pp.
  375--390, 2015.

\bibitem{galceran2017multipolicy}
E.~Galceran, A.~G. Cunningham, R.~M. Eustice, and E.~Olson, ``Multipolicy
  decision-making for autonomous driving via changepoint-based behavior
  prediction: Theory and experiment,'' \emph{Autonomous Robots}, vol.~41,
  no.~6, pp. 1367--1382, 2017.

\bibitem{pokorny2016topological}
F.~T. Pokorny, K.~Goldberg, and D.~Kragic, ``Topological trajectory clustering
  with relative persistent homology,'' in \emph{Robotics and Automation (ICRA),
  2016 IEEE International Conference on}.\hskip 1em plus 0.5em minus
  0.4em\relax IEEE, 2016, pp. 16--23.

\bibitem{li2018clustering}
S.~Li, W.~Wang, Z.~Mo, and D.~Zhao, ``Clustering of naturalistic driving
  encounters using unsupervised learning,'' \emph{arXiv preprint
  arXiv:1802.10214}, 2018.

\bibitem{bian2018survey}
J.~Bian, D.~Tian, Y.~Tang, and D.~Tao, ``A survey on trajectory clustering
  analysis,'' \emph{arXiv preprint arXiv:1802.06971}, 2018.

\bibitem{yao2017trajectory}
D.~Yao, C.~Zhang, Z.~Zhu, J.~Huang, and J.~Bi, ``Trajectory clustering via deep
  representation learning,'' in \emph{Neural Networks (IJCNN), 2017
  International Joint Conference on}.\hskip 1em plus 0.5em minus 0.4em\relax
  IEEE, 2017, pp. 3880--3887.

\bibitem{lake2015human}
B.~M. Lake, R.~Salakhutdinov, and J.~B. Tenenbaum, ``Human-level concept
  learning through probabilistic program induction,'' \emph{Science}, vol. 350,
  no. 6266, pp. 1332--1338, 2015.

\bibitem{nechyba1998stochastic}
M.~C. Nechyba and Y.~Xu, ``Stochastic similarity for validating human control
  strategy models,'' \emph{IEEE Transactions on Robotics and Automation},
  vol.~14, no.~3, pp. 437--451, 1998.

\bibitem{hochreiter1997long}
S.~Hochreiter and J.~Schmidhuber, ``Long short-term memory,'' \emph{Neural
  computation}, vol.~9, no.~8, pp. 1735--1780, 1997.

\bibitem{muller2007dynamic}
M.~M{\"u}ller, ``Dynamic time warping,'' \emph{Information retrieval for music
  and motion}, pp. 69--84, 2007.

\bibitem{taha2015efficient}
A.~A. Taha and A.~Hanbury, ``An efficient algorithm for calculating the exact
  hausdorff distance,'' \emph{IEEE transactions on pattern analysis and machine
  intelligence}, vol.~37, no.~11, pp. 2153--2163, 2015.

\bibitem{aronov2006frechet}
B.~Aronov, S.~Har-Peled, C.~Knauer, Y.~Wang, and C.~Wenk, ``Fr{\'e}chet
  distance for curves, revisited,'' in \emph{European Symposium on
  Algorithms}.\hskip 1em plus 0.5em minus 0.4em\relax Springer, 2006, pp.
  52--63.

\bibitem{liao2005clustering}
T.~W. Liao, ``Clustering of time series data: {A} survey,'' \emph{Pattern
  recognition}, vol.~38, no.~11, pp. 1857--1874, 2005.

\bibitem{xu2005survey}
R.~Xu and D.~Wunsch, ``Survey of clustering algorithms,'' \emph{IEEE
  Transactions on neural networks}, vol.~16, no.~3, pp. 645--678, 2005.

\bibitem{kanungo2002efficient}
T.~Kanungo, D.~M. Mount, N.~S. Netanyahu, C.~D. Piatko, R.~Silverman, and A.~Y.
  Wu, ``An efficient k-means clustering algorithm: Analysis and
  implementation,'' \emph{IEEE transactions on pattern analysis and machine
  intelligence}, vol.~24, no.~7, pp. 881--892, 2002.

\bibitem{wang2017much}
W.~Wang, C.~Liu, and D.~Zhao, ``How much data are enough? a statistical
  approach with case study on longitudinal driving behavior,'' \emph{IEEE
  Transactions on Intelligent Vehicles}, vol.~2, no.~2, pp. 85--98, 2017.

\end{thebibliography}
